%% file: bare_conf.tex
\documentclass[conference]{IEEEtran}
\usepackage{import}
\usepackage{example}

\begin{document}

\title{Going into Orbit: Massively Parallelizing Episodic Reinforcement Learning}

\author{\IEEEauthorblockN{Johann Bonneau}
\IEEEauthorblockA{Karlsruhe Institute of Technology\\Karlsruhe, Germany\\
Email: uivvm@student.kit.edu}
\and
\IEEEauthorblockN{Jan Oberst}
\IEEEauthorblockA{Karlsruhe Institute of Technology\\Karlsruhe, Germany\\
Email: uyvch@student.kit.edu}}

\maketitle

\begin{abstract}

The possibilities of robot control have multiplied across various domains through the application of deep reinforcement learning. To overcome safety and sampling efficiency issues, deep reinforcement learning models can be trained in a simulation environment, allowing for faster iteration cycles. This can be enhanced further by parallelizing the training process using GPUs. NVIDIA's open-source robot learning framework Orbit leverages this potential by wrapping tensor-based reinforcement learning libraries for high parallelism and building upon Isaac Sim for its simulations. We contribute a detailed description of the implementation of a benchmark reinforcement learning task, namely box pushing, using Orbit. Additionally, we benchmark the performance of our implementation in comparison to a CPU-based implementation and report the performance metrics. Finally, we tune the hyper parameters of our implementation and show that we can generate significantly more samples in the same amount of time by using Orbit.
\end{abstract}

\IEEEpeerreviewmaketitle

\section{Introduction}

Nowadays, robotics is a rapidly growing field. At the heart of it stands helping and assisting humans. Mainly, it aims at doing jobs under conditions that are hazardous or impossible for humans to operate under and replace humans in tasks that are tedious, monotonous, or disagreeable. Successful applications range from industrial tasks in warehouse management \cite{lee2018development} or automotive manufacturing \cite{bartovs2021overview} over medical tasks such as robot-assisted surgery \cite{enayati2016haptics} to social tasks in teaching \cite{benitti2012exploring} or geriatric care \cite{haubold2020introducing}. While in some fields, robots are controlled by humans or perform simple, directly programmable tasks, more sophisticated methods are required to enable robots to solve complex tasks autonomously.

To accomplish this, \gls{rl}, a subfield of machine learning, has proven to be highly successful \cite{zhao2020sim}. In numerous applications within robotics, tasks can naturally be modeled as \gls{rl} problems \cite{morales2021survey}. The concept behind \gls{rl} involves an agent interacting with an environment through actions and receiving feedback, referred to as reward \cite{sutton2018reinforcement}. The goal for the agent is to develop a policy based on which it decides which action to take in a specific scenario. Consequently, the agent learns to solve the task through trial-and-error by maximizing the cumulative reward, referred to as return. The basic approach is to provide feedback to the agent in a continuous fashion by updating its policy on each step \cite{sutton2018reinforcement}. While some tasks with a long span such as process-control go on continuously, others can be divided into closed sequences that end when a terminal state is reached \cite{sutton2018reinforcement}. These sequences of states and actions are called episodes and thus the learning process is called episodic \gls{rl}. Common examples of episodic \gls{rl} problems are the successful completion of games like Go, StarCraft, Atari games, or Tetris.

When applying \gls{rl} methods to robotic applications, often a large number of samples is required \cite{zhao2020sim}. Instead of performing the robot actions in a real environment, it is common practice to perform the training in a simulation first and transfer the result to a real setup \cite{zhao2020sim}. This not only speeds up the training process but also alleviates safety concerns with real robots \cite{zhao2020sim}. There are many existing simulation environments, e.g. popular ones such as Gazebo\footnote{Gazebo: \url{gazebosim.org}}, MuJoCo\footnote{MuJoCo: \url{mujoco.org}}, PyBullet\footnote{PyBullet: \url{pybullet.org}}, and Webots\footnote{Webots: \url{cyberbotics.com}}, all having different advantages and caveats \cite{korber2021comparing}. In contrast, Isaac Sim - a robotics simulator by NVIDIA - has been recognized for its particularly good integration with NVIDIA hardware. It is however still in its developmental stages. With Isaac Sim remaining closed-source, contributing to the simulator and creating a unified framework for research poses challenges for users \cite{mittal2024faq}. Therefore, NVIDIA recently released Orbit, a unified framework for interactive robot learning \cite{mittal2023orbit}. Its goal is to  allow the community to collaboratively push forward in developing benchmarks and robot learning systems. We contribute to this goal by implementing a benchmark learning environment, namely box pushing, laying the foundation for the evaluation of Orbit and the ecosystem surrounding it in the future. We additionally provide a black-box wrapper for the environment for use with movement primitives (MP). We successfully train agents in both scenarios and report the performance metrics. For the comparison in the evaluation, we choose Fancy Gym\footnote{Fancy Gym: \url{github.com/ALRhub/fancy_gym}} \cite{otto2024fancy_gym}, a framework for robotic reinforcement learning using MuJoCo for its simulations. Therefore, we closely follow the implementation of the box pushing environment in Fancy Gym to really highlight the difference in simulation capabilities of the highly parallelized Isaac Sim compared to the CPU-based MuJoCo. For our evaluation, we take the training results of the box pushing environment using Fancy Gym from \cite{otto2023deep}.

\section{Problem Statement}

In the following, we introduce Orbit and the ecosystem surrounding it. An overview of the components used in the experiments of this work are described subsequently. Finally, a detailed description of the created benchmark environment is provided.

\subsection{Structure of Orbit}

Orbit aims at allowing developers to efficiently create robot learning environments with fast and realistic physics in a straightforward way. It is powered by NVIDIA Isaac Sim, an extensible robot simulator which aims at providing physically accurate environments. Isaac Sim runs on top of the NVIDIA Omniverse platform, building on the \gls{openusd} file format and NVIDIA's raytracing technology NVIDIA RTX. Overall, Orbit offers fourteen robot articulations, three different physics-based sensors, twenty learning environments, and wrappers to five different learning frameworks -- namely RL Games, robomimic, RSL RL, Stable-Baselines3, and skrl \cite{mittal2023orbit}. Since both RL Games and RSL RL are optimized for GPUs, a training speed of 50,000-75,000 \gls{fps} with 2048 environments is reached by the authors of \cite{mittal2023orbit}.

The documentation of Orbit mentions an interesting feature of Isaac Sim: It allows developers to provide a seed and enable deterministic execution to make simulation runs reproducible \cite{mittal2024known}. However, we could not confirm this reproducibility in our experiments. As mentioned in \cite{mittal2024known}, the root cause for this could be the variance in GPU work scheduling which results in a varying order of execution and finally, leading to diverging environments.

\begin{figure}[!t]
\centering
\includegraphics[width=2.2in]{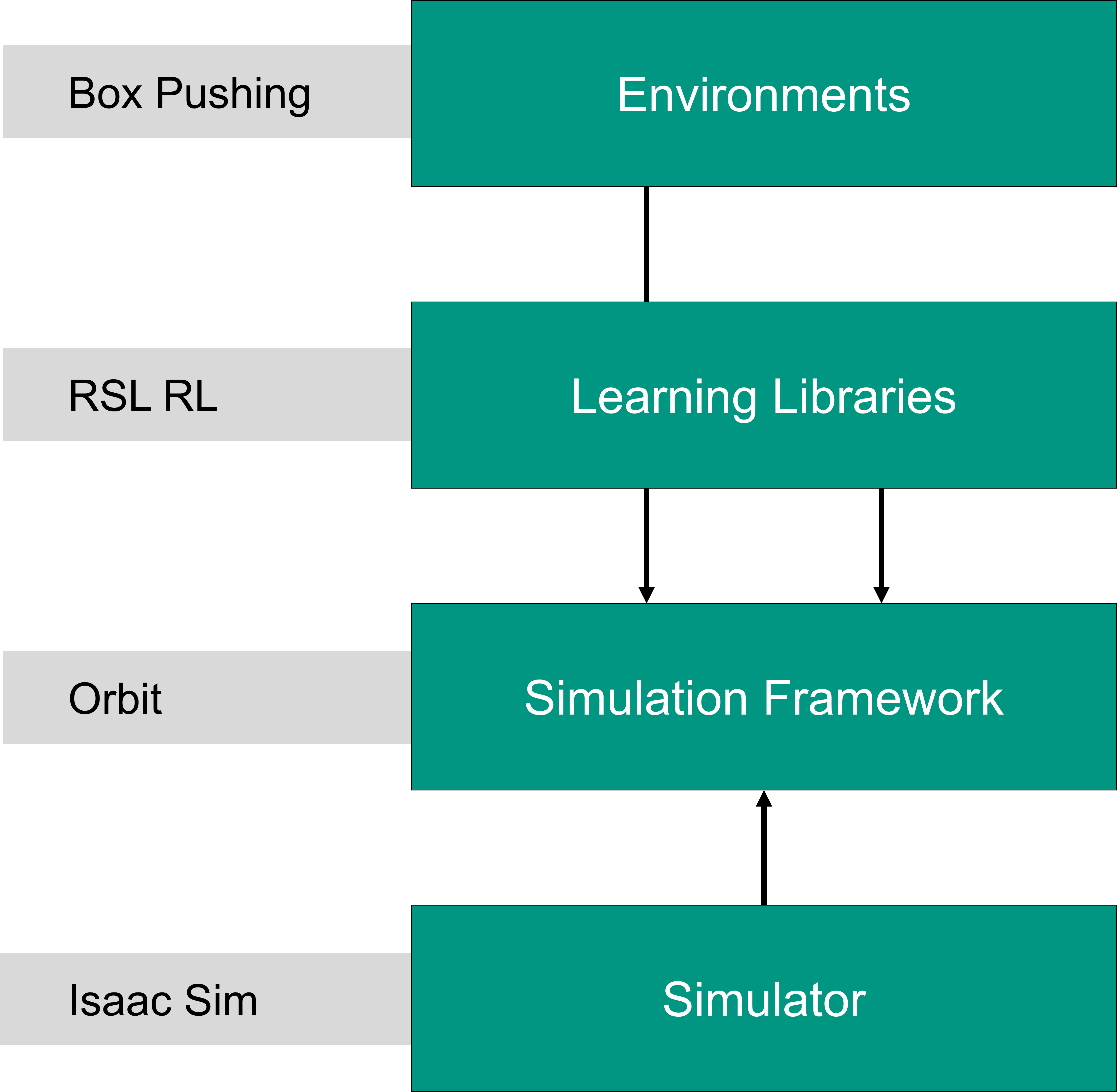}
\caption{Relationship of the components for the experiments in this study. Green: The simulation framework unifies the environments, learning libraries, and simulator to create a robot learning environment. Gray: Specific components used in this study.}
\label{fig:overview}
\end{figure}

The different components used for the experiments in this study are depicted in \cref{fig:overview}. The box pushing environment is defined, describing the physical surroundings and the actions that an agent can take. The exact setup is described in \cref{sub:benchmark-env}. Additionally, the machine learning library RSL RL is used to model and train the agent. Subsequently, Orbit serves as a middleman between the different components involved in the learning process: The agent's actions are sampled by RSL RL and the resulting state transitions are simulated in the defined box pushing environment using Isaac Sim. The generated samples are in turn used by RSL RL to improve the agent's policy. This process is repeated until the maximum number of training epochs is reached.

\subsection{Benchmark Environment Setup}
\label{sub:benchmark-env}

\begin{figure}[!t]
\centering
\includegraphics[width=2.4in]{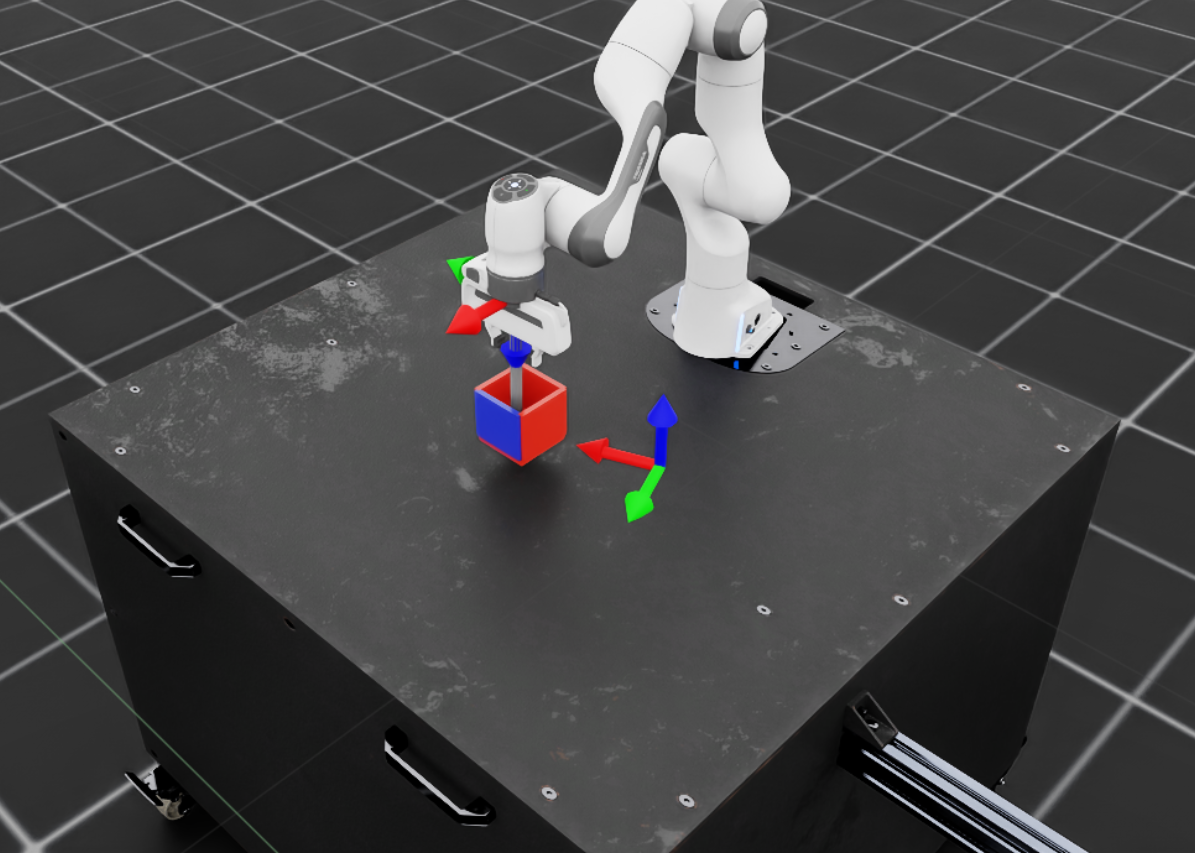}
\caption{Box pushing environment: The agent is supposed to push the box into a designated goal pose by moving the table-mounted Franka robot arm.}
\label{fig:sim-scene}
\end{figure}

We choose the box pushing task for our implementation because it is a good example for tasks that can be solved both in a step-based manner and using movement primitives. Besides that, the box pushing domain is a commonly used benchmark task for evaluating \gls{rl} methods in robotics (e.~g. \cite{otto2023mp3, plappert2018multi, andrychowicz2017hindsight}). For this task, the objective of the agent is to push the box from its initial position on the table to a target pose by actuating the robot arm's joints \cite{andrychowicz2017hindsight}. This target pose consists of a position on the table's surface and an orientation. The structure of the deduced environment can be divided into two parts: the physical arrangement of objects and their properties in the simulator and the learning specific components. 

\textbf{Physical arrangement}. The Franka robot arm is mounted to a table and its end effector is complemented by a push rod as depicted in \cref{fig:sim-scene}. The box consists of five quadratic plates with an edge length of ten centimeters resulting in a cube with an opening for the robot's rod to push the box. The box is placed on the table surface and the robot arm starts with the push rod inside the box. While the initial position and the orientation of the box as well as the robot arm's initial position are fixed, the box's target pose is generated randomly. 

\textbf{Reinforcement learning components.} We implement a reward function, termination criteria, and commands. The length of an episode is limited by two seconds, after which the environment is reset. As our goal was not to optimize for an optimal learning environment but rather to create a benchmark for comparability between different simulators, we closely followed the implementation used by \cite{otto2023deep}. The implementation details of the environment setup are described in the following section.

\begin{figure*}[!t]
\centering
\includegraphics[width=0.65\textwidth]{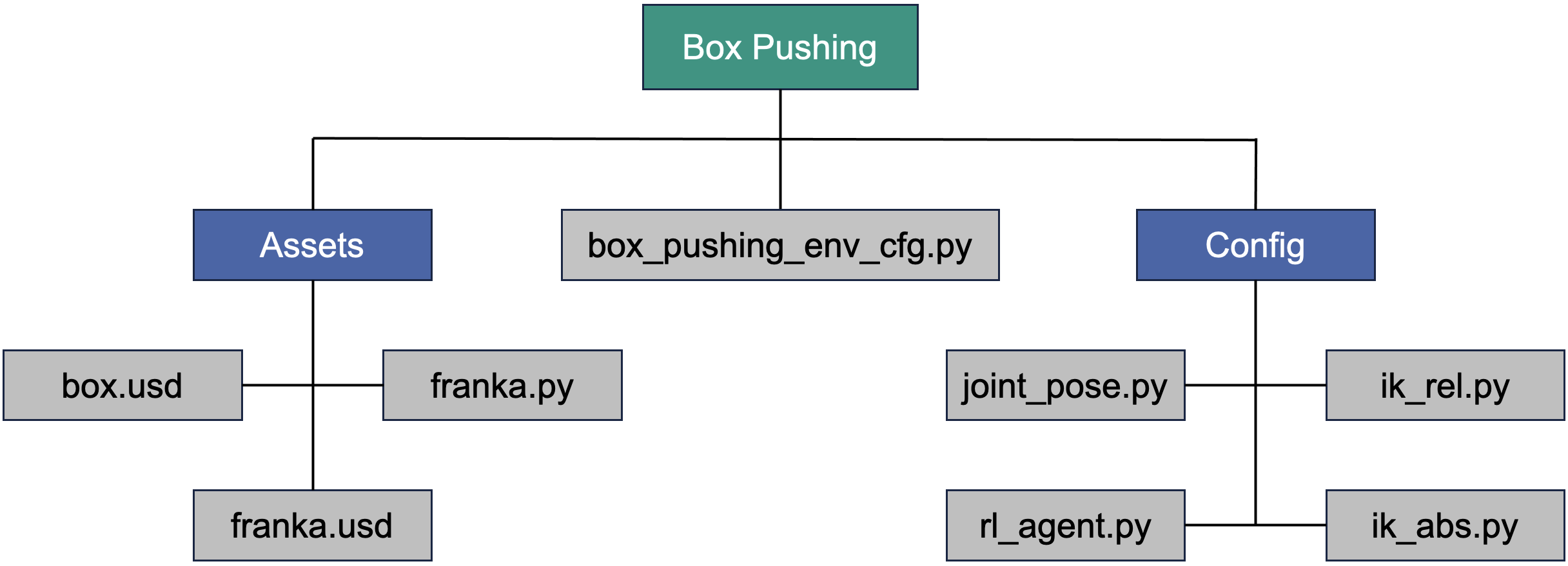}
\caption{Overview of the main file structure of the box pushing environment in Orbit (blue: folders, gray: files, Note: The MDP folder and the contained files are not represented as they are not relevant for the configuration of the environment).}
\label{fig:env-structure}
\end{figure*}

\section{Implementation}
The implementation is divided into two parts. First, we implement a step-based environment for the box pushing task. Second, we implement a black-box wrapper for the step-based box pushing environment in order to use movement primitives. In this chapter, we explain how to perform these implementations using the simulation framework Orbit.

\subsection{Step-Based Environment} \label{sec:step_env}

The structure of the step-based environment is shown in \cref{fig:env-structure}. In the \path{assets} folder, the files related to the assets of the scene and the configuration script for our robot can be found. This script configures the initial state of the robot, the parameters of the PD-controller and the limits of the actuator groups. In our environment, we use implicit actuators. The main part of the configuration of the environment takes place in \path{box_pushing_env_cfg.py}. This script is extended with additional scripts in the \path{Config} folder which configure whether to perform the task in the joint space or in the task space and finalize the configuration.

The configuration consists of implementing a configuration class for the components of the environment in the \path{box_pushing_env_cfg.py} script. These configuration classes are then instantiated in the main configuration class. A code snippet of the configuration script is shown in the appendix in Listing~\ref{lst:env-cfg}. Configuring the components is simple as it consists of defining variables containing instances of configuration classes. The implementation of a component of the environment is explained in \cref{sec:base-env} for the scene design.

In Orbit, the configuration of the step-based \gls{rl} environment can be divided in two parts: configuration of the base environment which is then supplemented to support \gls{rl}.

\subsubsection{Creating a Base Environment} \label{sec:base-env}
\hfill\\
\textbf{Scene definition.} The first step to creating the base environment is to design the scene. Therefore, we recreate the experiment setup in the simulator. We use already existing assets for the table, the ground plane and the light. The \gls{openusd} file for assets provided by Isaac Sim are stored in a Nucleus database. For this scene, we use a customized \gls{openusd} file for the 7-DoF Franka robotic arm with a cylinder placed in the gripper acting as a push rod. Finally, we recreated the box used by Fancy Gym. The box files in Fancy Gym are stored in the MJCF format but we encountered difficulties using the MJCF converter of Isaac Sim. We therefore recreated the box with a CAD software and then imported the STEP file in Isaac Sim in order to save it to a \gls{openusd} file. The custom \gls{openusd} files we created are stored locally in the \path{assets} folder. In the appendix in Listing~\ref{lst:env-cfg}, we show how the scene is configured in Orbit. We simply add variables containing the configuration classes of the assets in the scene. The variables for the robot and the manipulated object are left empty to enable further customization without having to define multiple scenes.

\textbf{Actions.} After creating the scene, we define the actions of the robot. In other terms, we define what type of controller is used to input actions to the robot. Orbit allows us to define different controllers for different sets of joints. We use a joint position controller for the arm and hand joints. No controller for the gripper joints is assigned as they are not used in our experiment.

\textbf{Observation.} We then define the states of the environment that can be observed by an agent. In Orbit, we can define multiple observation groups that serve different purposes. We only use the default group named "policy" which is used to observe the scene after an episode is finished. The observations relevant to our environment are:

\begin{itemize}
    \item the joint positions,
    \item the joint velocities,
    \item the object pose,
    \item the target pose and
    \item the last action performed.
\end{itemize}

\begin{figure*}[!h]
    \centering
	\resizebox{0.75\textwidth}{!}{%
		\input{tikz/BBRL_overview}
	}%
	\caption{Figure taken from \cite{otto2023deep}. Overview of the proposed Black-Box Reinforcement Learning (BBRL) framework. The normal distributed policy predicts, given the context ${c}$, the parameters of a movement primitive that translates to a desired trajectory $\tau^d$. A trajectory tracking controller $f$ generates low-level actions $a_t$ given the current state $s_t$ and the desired state $s_t^d$ from the trajectory $\tau^d$.}
	\label{fig:schema}
\end{figure*}
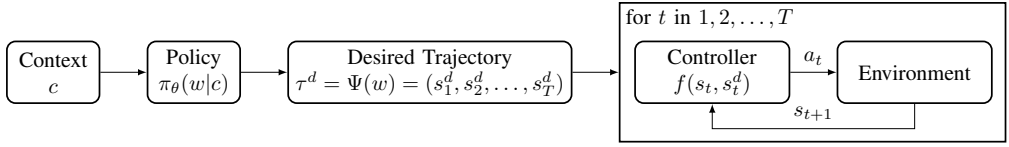

\textbf{Randomization.} To finalize the base environment, we configure the randomization. At the start, the push rod has to be inside the box. Randomizing the initial box position would therefore require the computation of the joint angles with inverse kinematics at the start of every episode. We therefore fix the initial box position and only randomize the target pose. A random initial box position could however be subject to future work.

\subsubsection{Creating a RL Environment}
\hfill\\
To allow for the environment to be used for \gls{rl} tasks, it is complemented with a reward function, termination criteria, commands, and curriculums. 

\textbf{Reward.} The reward function is implemented by defining reward terms. The terms use a function to compute a return given the observation. Given the reward terms $R_{0}, \dots, R_{n}$ and their weights $w_{0}, \dots, w_{n}$, the total reward $R$ is:
$$R = \sum_{i=0}^{n} w_i \cdot R_i$$

We implement the dense reward from \cite{otto2023deep}. The reward terms are:

\begin{itemize}
    \item $R_{goal}$: the distance between the box position and the target position.
    \item $R_{rot}$: the angular distance between the box orientation and the target orientation around the yaw-axis.
    \item $R_{en}$: the energy cost.
    \item $R_{lim}$: the sum of the values exceeding the joint position and velocity limits
    \item $R_{rod}$: the distance between the tip of the push rod and the box.
    \item $R_{rod\_rot}$: the distance between the current orientation and the target orientation of the rod. The goal of this term is to keep the push rod pointing down.
\end{itemize}
The total reward $R$ is
$$R = - R_{rod} - R_{rod\_rot} - \num{5e-4}R_{en} - R_{lim} - 2 R_{rot} - 3.5 R_{goal}.$$

\textbf{Termination Criteria.} An episode can either be terminated by a timeout or be truncated by a condition that becomes true. The timeout is set to 2 seconds. Additionally, we implement the success condition from \cite{otto2023deep}. An episode "is considered successfully solved if the position distance $\leq 0.05$ \unit{\metre} and the orientation error $\leq 0.5$ \unit{rad}". This condition is however not used for truncation of an episode. Since the box's velocity is not regarded by the condition, the agent would launch the box to the target pose as quickly as possible. As our goal is for the agent to move the box to the target pose in a controlled way, this behavior is undesired and no truncation is performed.

\textbf{Commands.} The commands are used for goal condition tasks. They are resampled at the end of every episode and can also be resampled after a specified time period. We implemented a command that samples a new box target pose. As in \cite{otto2023deep}, the command uniformly samples the pose between $[0.3;0.6]$ for the X-axis, $[-0.45;0.45]$ for the Y-axis and $[0;2\pi]$ for the yaw-axis.

\textbf{Curriculum.} Orbit allows to implement curriculum learning by implementing curriculum terms. The box pushing task does not require curriculum learning. 

\subsection{Black-Box Environment}

Tasks involving physical interaction oftentimes can be divided into different action phases \cite{stulp2012reinforcement}. By solving separate subtasks -- so called movement primitives -- individually, the search space is reduced drastically making the training process both more efficient and more robust. Therefore, we wrap the existing step-based environment to allow for episodic \gls{rl} using movement primitives.

\subsubsection{Black-Box Reinforcement Learning} \label{sec:erl}

\begin{figure*}[!t]
\centering
\includegraphics[width=0.75\textwidth]{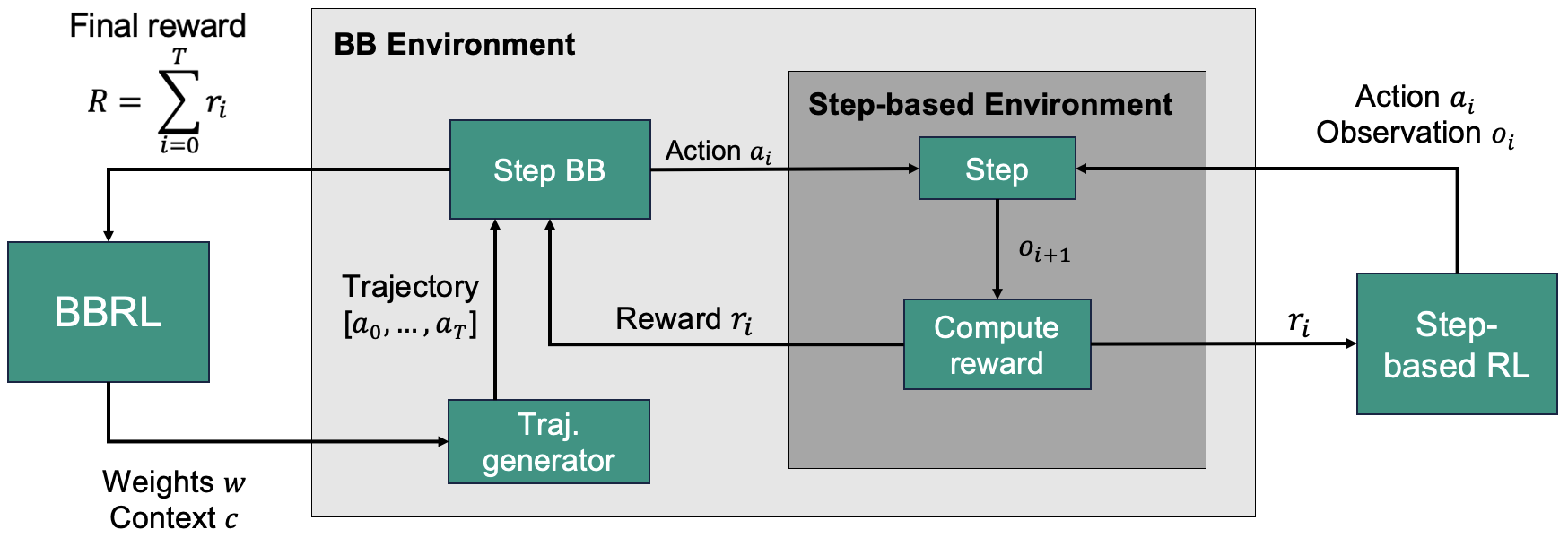}
\caption{Overview of the interaction of step-based and black-box reinforcement learning (BBRL) agents with their environment. Right: Step-based reinforcement learning agent performing a step. Left: Black box reinforcement learning agent performing a step.}
\label{fig:overview-rl}
\end{figure*}

The goal of BBRL is to learn a policy $\pi_\theta(w|c)$ where $c$ is the context and $w$ is a weight vector used to get the reference trajectory $\Psi(w)$ for the episode. To get the context, we apply a context mask to the initial observation of the episode as follows: $$c = m * o_0 $$ with $m \in \{0,1\}^n$ being the mask and $o_0$ being the observation vector of dimension $n$. The weight vector is used by a trajectory generator $\Psi$ to get the reference trajectory $\Psi(w) = [s_t]_{t=0:T}$. In our work, we use Probabilistic Movement Primitives (ProMP) \cite{paraschos2013probabilistic} to generate the trajectories.

As depicted in \cref{fig:overview-rl}, the BBRL environment acts as a black-box by wrapping around the step-based environment. During an episode, the weight vector $w$ and the context $c$ are taken as input, based on which the reference trajectory is generated and subsequently executed step-by-step to return the final reward. At each step the rewards are collected and then used to compute the final reward of the episode (e.g. the sum of the collected rewards).

\subsubsection{MP-based environment}

With the step-based environment being implemented as explained in \cref{sec:step_env}, a wrapper around the step-based environment is implemented to provide a BBRL environment. The goal of the wrapper is to apply the context mask to the observation vector and to override the step function. The step function executes a trajectory using a given weight vector and returns the reward. The initialization of the environment follows these steps. First, we initialize the step-based environment as usual. Then, we gather the configurations of the different elements of the BBRL environment e.g. the parameters of the trajectory generator or the length of a trajectory. We then initialize these components to finally return the BBRL environment implementing the same functions as the step-based environment. This allows us to use both environments interchangeably with the same training script by just changing the used environment and the algorithm hyper parameters.

\textbf{Step Function.} A step in the BBRL environment represents an entire episode in the step-based environment. The first task is to get the reference trajectory. Therefore, we use the trajectory generator that was initialized with its parameters. The trajectory generator takes as input:

\begin{itemize}
    \item the start step $t_s$ (in our case $t_s=0$),
    \item the current joint angles,
    \item the current joint velocities,
    \item the length of the trajectory $T$ (in our case $T = 100$~steps),
\end{itemize}

and returns a list of actions to be executed. The trajectory is then executed using the step-based environment. At each step $t$ of the trajectory, the current reward $r_t$ is computed and collected. After the trajectory is executed, we compute the final reward $R$ which is the sum of all collected rewards: $$R = \sum_{t=t_s}^{T} r_t\text{.}$$

\textbf{Observation Function.} As explained in \cref{sec:erl}, the observation is computed by taking the observation of the step-based environment and applying the context mask. The returned vector contains only the values filtered by the context mask.

To realize this, we used the wrapper in Fancy Gym that was already implemented and modified it to execute operations on a GPU with PyTorch. We extended this wrapper to use the configurations and context mask for our task. We also override the functions that are responsible for returning the current joint angles and velocities in order to get these values from Isaac Sim.

\section{Results \& Evaluation}

The evaluations of both the step-based and the MP-based implementation are performed in two steps. First, we run the training applying the hyper parameters used by \cite{otto2023deep} to our setting for a benchmark and compare their performance in terms of success rate per number of environment interactions. Second, we adapt the hyper parameters to better fit the capabilities of Orbit and demonstrate the strength of its high parallelism and good integration with NVIDIA hardware.

\begin{table}[ht]
\centering
\begin{adjustbox}{max width=\textwidth}
\begin{tabular}{ll}
\textbf{Component}  &\textbf{Specification} \\ 
\hline
\multicolumn{2}{l}{} \\
Processor & AMD Ryzen 9 7900X \\
GPU & NVIDIA GeForce RTX 4080 16GB \\
RAM & 64GB \\
\multicolumn{2}{l}{} \\ 
\hline
\multicolumn{2}{l}{} \\
OS Version & Ubuntu 22.04.4 LTS \\
Python Version & 3.10 \\
PyTorch & 2.1.0 \\
CUDA & 11.7 \\
\multicolumn{2}{l}{} \\ 
\hline
\\
\end{tabular}
\end{adjustbox}
\caption{Experiment Setup Specifications}
\label{table-experiment-setup}
\end{table}

For all experiments, the hardware and software versions displayed in \cref{table-experiment-setup} are used. Per experiment, ten trials are performed.

\subsection{Benchmark: Orbit vs. Fancy Gym} \label{sec:orbit_vs_fancy-gym}

For the first part of the evaluation, the comparison between Fancy Gym and Orbit for the step-based and the BBRL environments is covered. In order to produce comparable results, we set the number of environments to 160 and the number of steps per epoch to 100 in Orbit to match the 16000 samples performed at each learning iteration for the training of the Fancy Gym step-based environment. For the BBRL environments, the number of steps per epoch is set to 1. The full set of hyper parameters can be found in the appendix.

\begin{figure*}[t]
    \centering
	\resizebox{\textwidth}{!}{
		\input{tikz/joints_comp}
	}
    \resizebox{0.8\textwidth}{!}{
        \input{tikz/legend_act_traj}
    }%
	\caption{The reference trajectory and the actual trajectories from Fancy Gym and Orbit for each joint of the Franka arm.}
    \label{fig:act_traj}
\end{figure*}
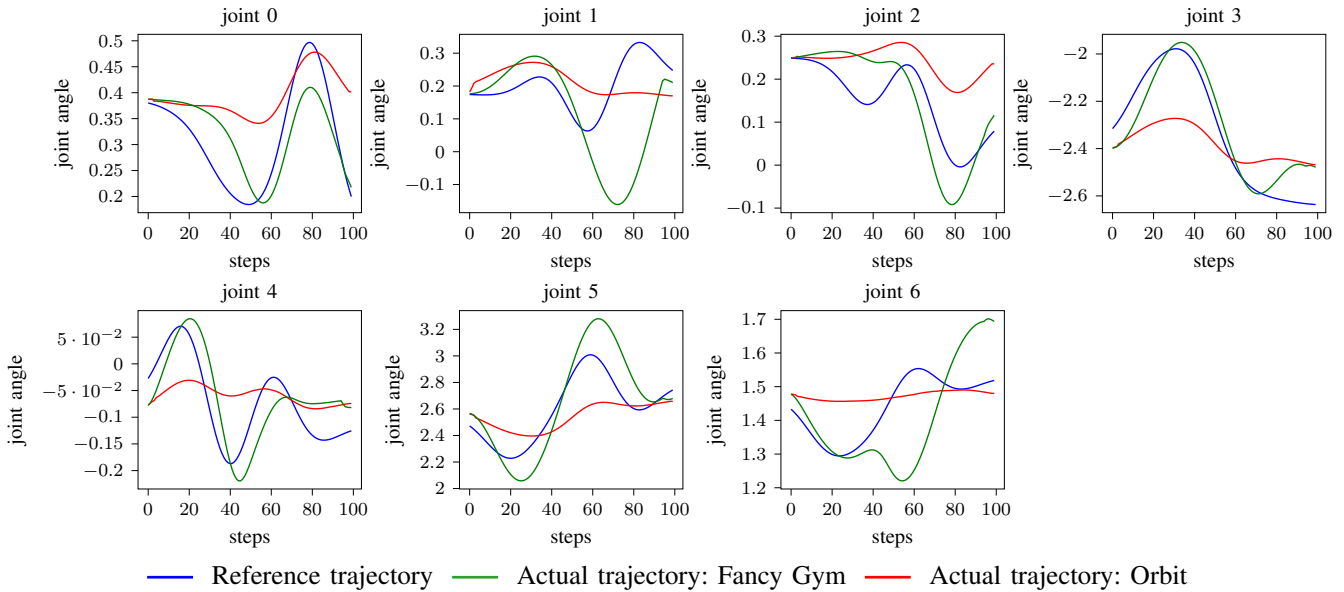

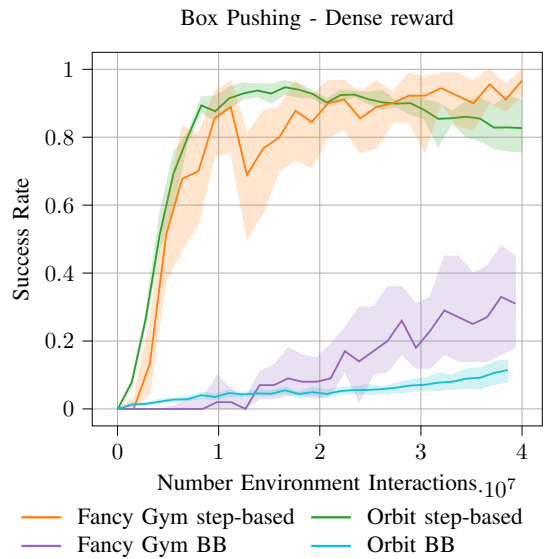
\begin{figure}[t]
    \centering
    \begin{minipage}{0.4\textwidth}
    \resizebox{\textwidth}{!}{\input{tikz/box_pushing_FG_HP}}%
    \end{minipage}%
    \\
    \resizebox{0.4\textwidth}{!}{
    \input{tikz/legend_eval_FG_HP}
    }%
    \caption{The success rate of the training experiments for the 4 environments relative to the number of environment interactions.}
    \label{fig:fg-hp-eval}
\end{figure}

\textbf{Step-based benchmark.}
Looking at the results of the training in \cref{fig:fg-hp-eval}, similar success rates for the step-based environments of Fancy Gym and Orbit can be observed. However, some differences are worth noting: First, the training of the step-based environment using Orbit achieves a higher success rate at approximately $1.2\cdot10^7$ environment interactions, decreasing from that point onwards. In contrast, the success rate of the Fancy Gym step-based environment does not stop increasing to finally surpass the success rate of the Orbit step-based environment. A possible explanation for these differences could be that we use different learning libraries for the training. 

We can conclude that Orbit can achieve a similar success rate compared to Fancy Gym when training the box pushing task using step-based \gls{rl}.

\textbf{MP-based benchmark.}
The results of the training using the BBRL environments are also depicted in \cref{fig:fg-hp-eval}. A difference in the success rate during the training can be observed. While the environment of Fancy Gym reaches a success rate of around 0.36 at $4\cdot10^7$ environment interactions the environment of Orbit is only able to achieve a success rate of 0.1. 

This difference can possibly be explained by the difference in the behavior of the robotic arm in MuJoCo and Isaac Sim. Given the same weight vector $w$, the reference trajectories of Orbit and Fancy Gym are approximately the same: $$\Psi_{Orbit}(w) \approx \Psi_{FancyGym}(w).$$ The difference between both simulators lies in the actual trajectories. The actual trajectory is the trajectory that is executed in the simulators based on the given reference trajectory. The actual trajectory is composed of the joint angles of the robot after executing a step with an action from the reference trajectory. \cref{fig:act_traj} shows the joint angle values of the reference trajectory and the actual trajectories of Fancy Gym and Orbit. We can note that the actual trajectories are not the same. That shows a difference in behavior of the robots in the simulator that could explain the difference in the success rate of the black-box environments of Fancy Gym and Orbit.

\subsection{Tuning Orbit}

To demonstrate the capabilities of Orbit, we performed the training of the box pushing environment using the hyper parameters of a similar task that is already implemented in Orbit. The hyper parameters are presented in \cref{tab:boxpushing-HP_2}. A complete overview of the hyper parameters is provided in the appendix. The biggest strength of Orbit is that it is able to run multiple environments on one simulation instance on one GPU. In this case, the training is performed by using 4096 environments in parallel. 

\begin{table}[ht]
\centering
\begin{adjustbox}{width=.45\textwidth}
\begin{tabular}{lcc}
                                 & Previous BBRL-HP          & Tuned Orbit BBRL-HP    \\ 
\hline
\multicolumn{3}{l}{}                                                                                                  \\
number samples                   & 160        & 4096     \\
\# parallel env. in Orbit     & 160         &4096       \\
\multicolumn{3}{l}{}                                                                                                  \\ 
\hline
\multicolumn{3}{l}{}                                                                                                  \\
hidden layers                    & {[}256, 256] & {[}256, 128, 64]  \\
hidden layers critic             & {[}256, 256] & {[}256, 128, 64]    \\
hidden activation                & tanh         & elu          \\     
\end{tabular}
\end{adjustbox}
\caption{Comparison of hyper parameters. First column: HP previously used in \cref{sec:orbit_vs_fancy-gym} for the benchmark with Fancy Gym. Second column: HP used to show the capabilities of Orbit.}
\label{tab:boxpushing-HP_2}
\end{table}

\begin{figure*}[!t]
\centering
\subfloat[The success rate during the training for both sets of hyper parameters relative to the training time]{
\includegraphics[width=2.5in]{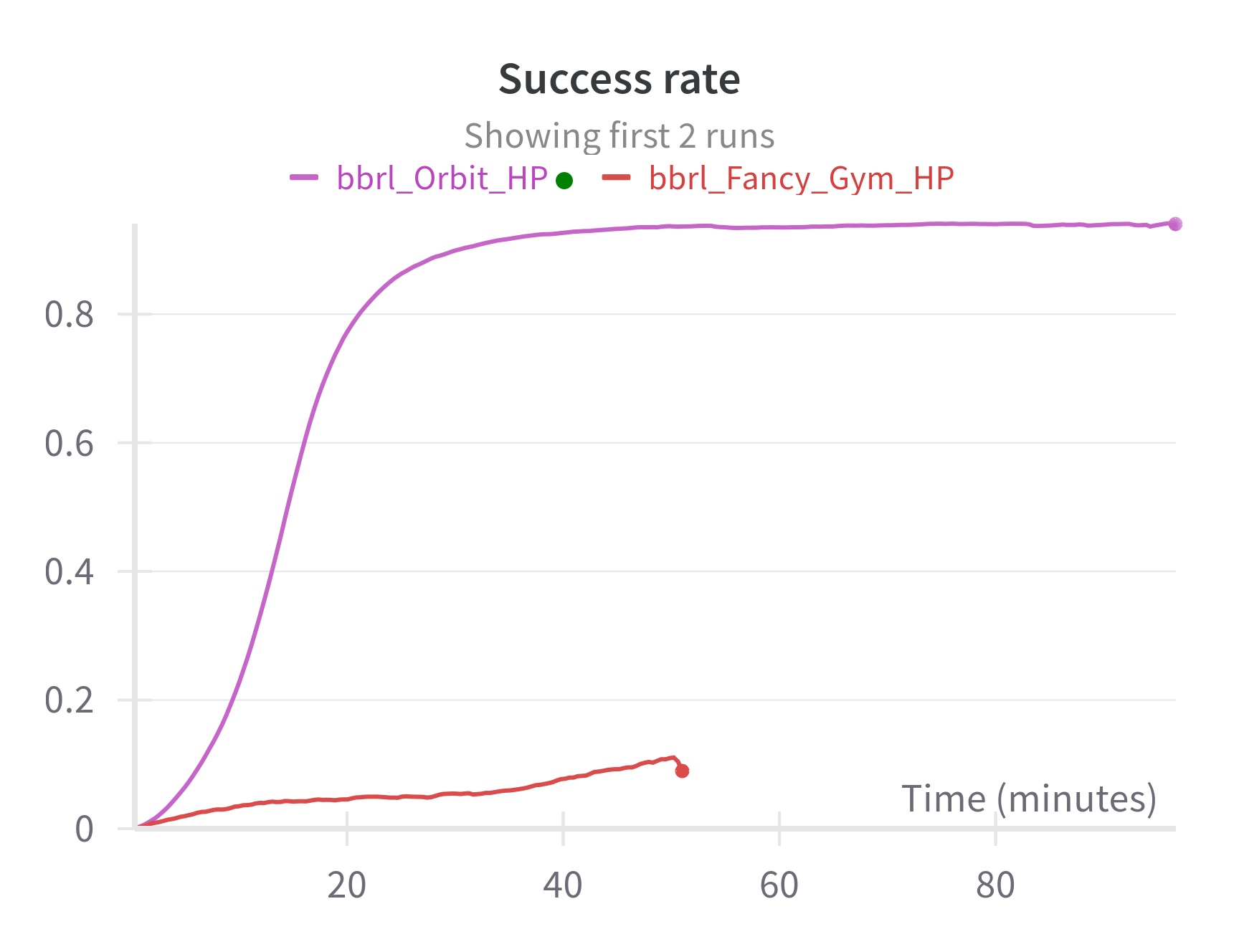}%
\label{fig:bbrl_orbit_tuning_success}
}
\hfil
\subfloat[The amount of environment interactions during the training for both sets of hyper parameters relative to the training time]{
\includegraphics[width=2.5in]{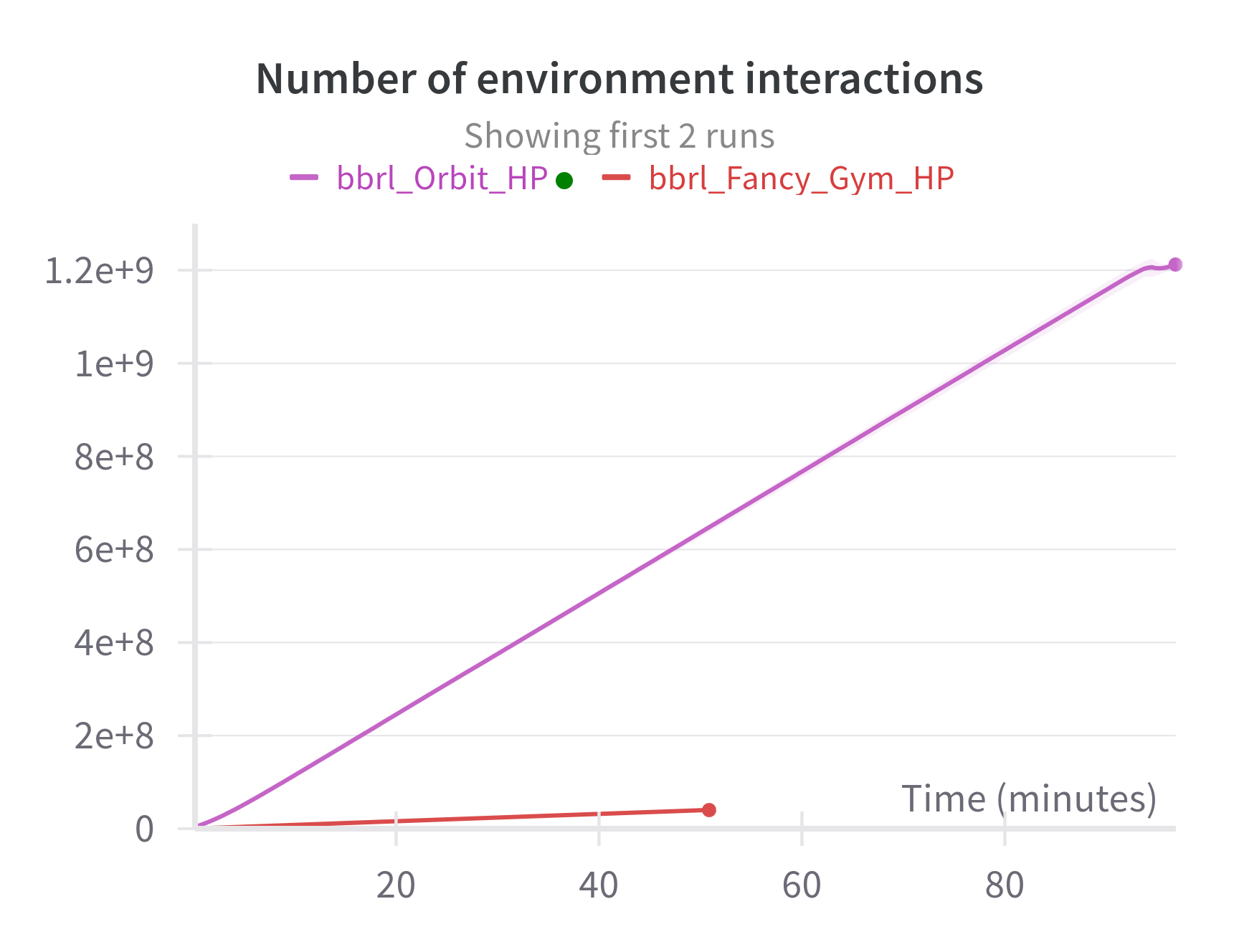}%
\label{fig:bbrl_orbit_tuning_time}
}
\caption{Comparing the success rate and the number of environment interactions for the two set of hyper parameters}
\label{fig:bbrl_orbit_tuning}
\end{figure*}

This parallelism capability is particularly notable when looking at the training time and the number of interactions that were performed in that time. On \cref{fig:bbrl_orbit_tuning_success}, we can see that the success rate of the BBRL environment using 4096 environments reaches values close to 0.93 after 50 minutes of training time compared to 0.1 for 160 environments. Thus, with the same training time, we can achieve a higher success rate. This is because with 4096 parallel environments, Orbit is able to gather more samples in the same amount of time as shown in \cref{fig:bbrl_orbit_tuning_time}.

To conclude, by increasing the number of environments used for the training, we can achieve a higher success rate for BBRL. This showcases the capabilities of Orbit and Isaac Sim when dealing with RL tasks.

\section{Conclusion \& Outlook}

This study presented a detailed exploration of the application of reinforcement learning in the rapidly growing field of robotics by developing and evaluating a benchmark learning environment within NVIDIA's Orbit framework. Through the implementation of a box pushing task using a Franka robot arm in a simulated environment, this work demonstrated the potential of reinforcement learning in enabling robots to perform complex tasks autonomously. By leveraging NVIDIA's Isaac Sim for the simulation, and Orbit for the integration and execution of reinforcement learning algorithms, we successfully established a benchmark for evaluating the efficiency and effectiveness of reinforcement learning approaches in robotic applications.

Our findings reveal that using Orbit for the application of reinforcement learning to robotics holds significant promise for advancing the capabilities of robots. Overall, the importance of simulation in the training process is underscored for enhancing the learning efficiency and accelerating iteration cycles.

Despite the promising results, our experiments also highlighted challenges related to reproducibility and the variance in behaviors between different simulators, suggesting areas for further refinement. The introduction of the Orbit framework represents a substantial step forward in the collaborative development of robot learning systems and also opens up new avenues for research.

Future research should focus on overcoming these challenges. On the one hand, the implemented environment could be expanded to create a more sophisticated simulation environment that can more accurately mirror the complexities of the real world. For example, a random initial pose of the robot and the box could be modeled in the future. On the other hand, expanding the scope of benchmark tasks and exploring other benchmark environments to further investigate the capabilities of different simulators could be subject to future research. More sophisticated benchmarks and evaluation of Orbit compared to other simulation frameworks is highly encouraged to carve out the strengths and limitations of Orbit. Additionally, other reinforcement learning techniques could be explored such as Trust Region Project Layers. Ultimately, the continued evolution of reinforcement learning in robotics promises to significantly expand the boundaries of what is achievable, paving the way for robots to become even more integral and beneficial components of human society.

\section*{Acknowledgment}
The authors would like to thank their supervisors, Tai Hoang and Ge Li, for their valuable guidance, support, and insightful feedback throughout the course of this project. Their expertise was inspiring and significantly contributed to the successful completion of this report.

\bibliographystyle{plain} % We choose the "plain" reference style
\bibliography{bibliography} 

\include{appendix}

\end{document}

%% file: tikz/BBRL_overview.tex
\tikzset{node style/.style={draw, thick, rounded corners, minimum height=1cm, minimum width=1.5cm, align=center}}

\begin{tikzpicture}
    % grid for easier spacing
    % \draw[help lines,step=5mm,gray!20] (-0.5,-5) grid (25,5);
    
    % nodes 
    \node[node style] (c) at (0,0) {Context\\${c}$};  
    \node[node style, right = 0.75cm of c] (pi) {Policy\\$\pi_{{\theta}}({w} \vert {c})$};
    % \node[node style, right = 0.5cm of pi, minimum width=2.5cm] (w) {$\bm{w} \sim \mathcal{N}(\tilde{\mu},\tilde{\Sigma})$};
    \node[node style, right = 0.75cm of pi, minimum width=2.5cm] (tau) {Desired Trajectory\\${\tau}^d = \Psi({w}) = ({s}_1^d, {s}_2^d, \ldots, {s}_T^d)$};
    
    % ENV loop on top of each other
    % \node[node style, above right = 0.01cm and 1cm of tau, minimum width=2.5cm, align=center] (controller) {Controller\\$p_\tau(s_t)$};
    % \node[node style, below right = 0.01cm and 1cm of tau, minimum width=2.5cm] (env) {Environment};
    % \node[draw, thick, minimum height=3.5cm, minimum width=4cm, right = 0.5cm of tau] (loop) {};
    
    % ENV behind of each other
    \node[node style, right = of tau, minimum width=2.5cm] (controller) {Controller\\$f({s}_t, {s}_t^d)$};
    \node[node style, right = 0.75cm of controller, minimum width=2.5cm] (env) {Environment};
    \node[draw, thick, minimum height=2.25cm, minimum width=6.25cm, right = 0.75cm of tau] (loop) {};
    \node[anchor=north west,inner sep=3pt] at (loop.north west) {for $t$ in $1,2,\ldots, T$};
    
    % arrows
    \draw[-latex]
        (c) edge (pi) 
        % (pi) edge (w)
        % (w) edge (tau)
        (pi) edge (tau)
        (tau) edge (loop)
        (controller) edge node[node distance=2.5cm, auto] {$a_t$} (env)
        % (env.east) -- +(0.5,0.0) |- node[above, pos=0.25] {$s_{t+1}$} (controller.east)
        (env.south) -- +(0.0,-0.4) -| node[above, pos=0.25] {$s_{t+1}$} (controller.south)
        ;

\end{tikzpicture}

%% file: tikz/joints_comp.tex
% This file was created with tikzplotlib v0.10.1.
\begin{tikzpicture}

\definecolor{darkgray176}{RGB}{176,176,176}
\definecolor{green01270}{RGB}{0,127,0}

\begin{groupplot}[group style={
group size=4 by 2,
horizontal sep=1.5cm,
vertical sep=1.5cm,
},
footnotesize,
xlabel=steps,
ylabel=joint angle,
]
\nextgroupplot[
tick align=outside,
tick pos=left,
title={joint 0},
x grid style={darkgray176},
xmin=-4.95, xmax=103.95,
xtick style={color=black},
y grid style={darkgray176},
ymin=0.168768106400967, ymax=0.512497813999653,
ytick style={color=black},
]
\addplot [semithick, blue]
table {%
0 0.380116850137711
1 0.379163950681686
2 0.378102481365204
3 0.376924842596054
4 0.37562370300293
5 0.374191761016846
6 0.372621893882751
7 0.37090677022934
8 0.369038581848145
9 0.367008864879608
10 0.36480787396431
11 0.362424284219742
12 0.359844923019409
13 0.35705441236496
14 0.354035407304764
15 0.350768566131592
16 0.347233444452286
17 0.34340900182724
18 0.339275002479553
19 0.334813445806503
20 0.330010175704956
21 0.324856698513031
22 0.319352000951767
23 0.313504040241241
24 0.307331144809723
25 0.300862610340118
26 0.294138789176941
27 0.287210285663605
28 0.280136346817017
29 0.272982507944107
30 0.265817761421204
31 0.258711338043213
32 0.251729607582092
33 0.244933187961578
34 0.238374873995781
35 0.232098340988159
36 0.226137518882751
37 0.220517382025719
38 0.215254783630371
39 0.210360527038574
40 0.205841675400734
41 0.201703980565071
42 0.197954535484314
43 0.194604068994522
44 0.191669151186943
45 0.189173966646194
46 0.187151491641998
47 0.185644149780273
48 0.184704005718231
49 0.184392184019089
50 0.184777960181236
51 0.185937374830246
52 0.187951326370239
53 0.190903469920158
54 0.194877743721008
55 0.199955731630325
56 0.206213429570198
57 0.21371753513813
58 0.222521707415581
59 0.232661426067352
60 0.244149401783943
61 0.256970047950745
62 0.271074146032333
63 0.286374121904373
64 0.302740573883057
65 0.319999635219574
66 0.337934225797653
67 0.356286108493805
68 0.374762147665024
69 0.393042147159576
70 0.410789549350739
71 0.42766273021698
72 0.44332692027092
73 0.457465529441833
74 0.469789981842041
75 0.480047255754471
76 0.488026171922684
77 0.493560791015625
78 0.496533036231995
79 0.496873736381531
80 0.494563341140747
81 0.4896320104599
82 0.482159644365311
83 0.472274959087372
84 0.46015453338623
85 0.446019470691681
86 0.430131375789642
87 0.412785172462463
88 0.394300818443298
89 0.375012099742889
90 0.35525518655777
91 0.335356563329697
92 0.315620899200439
93 0.296321719884872
94 0.277693688869476
95 0.259927898645401
96 0.243170350790024
97 0.227522805333138
98 0.213045656681061
99 0.199762940406799
};
\addplot [semithick, red]
table {%
0 0.387263476848602
1 0.387568950653076
2 0.387678146362305
3 0.383484333753586
4 0.38418036699295
5 0.383085310459137
6 0.383016467094421
7 0.382159292697906
8 0.381733566522598
9 0.38109764456749
10 0.380537658929825
11 0.379920393228531
12 0.379338353872299
13 0.378760397434235
14 0.378211796283722
15 0.377688497304916
16 0.377200484275818
17 0.376755684614182
18 0.376359552145004
19 0.376008808612823
20 0.375703513622284
21 0.37544196844101
22 0.375221312046051
23 0.375054091215134
24 0.374953985214233
25 0.374871253967285
26 0.374791771173477
27 0.374719679355621
28 0.374605119228363
29 0.374468773603439
30 0.374291300773621
31 0.374045878648758
32 0.373701095581055
33 0.373236656188965
34 0.372632592916489
35 0.371871709823608
36 0.37093859910965
37 0.369821131229401
38 0.368512183427811
39 0.367009550333023
40 0.365315556526184
41 0.363439172506332
42 0.361398458480835
43 0.359216809272766
44 0.356929570436478
45 0.354593694210052
46 0.352245390415192
47 0.349965572357178
48 0.347769260406494
49 0.345763862133026
50 0.344006538391113
51 0.342582732439041
52 0.341630309820175
53 0.341158241033554
54 0.341008871793747
55 0.34147784113884
56 0.34236952662468
57 0.344313353300095
58 0.346796274185181
59 0.350073128938675
60 0.354029417037964
61 0.358688861131668
62 0.364040076732635
63 0.369993925094604
64 0.3766750395298
65 0.383806616067886
66 0.391544997692108
67 0.399504572153091
68 0.407804429531097
69 0.416160941123962
70 0.424363166093826
71 0.432488679885864
72 0.44026717543602
73 0.447584718465805
74 0.454158216714859
75 0.460339307785034
76 0.465609937906265
77 0.470080584287643
78 0.473530292510986
79 0.476064771413803
80 0.477629780769348
81 0.478226095438004
82 0.477851510047913
83 0.476541429758072
84 0.474344313144684
85 0.471330136060715
86 0.467581331729889
87 0.463194161653519
88 0.458280146121979
89 0.452965259552002
90 0.447372078895569
91 0.441537320613861
92 0.435590952634811
93 0.429613023996353
94 0.423687636852264
95 0.417893618345261
96 0.412278890609741
97 0.406890481710434
98 0.401762008666992
99 0.401762008666992
};
\addplot [semithick, green01270]
table {%
0 0.387148633969617
1 0.387124498974768
2 0.386907937831486
3 0.386619545095916
4 0.386302277808423
5 0.385985292388891
6 0.385680004591118
7 0.38538718928101
8 0.385100522088363
9 0.384809361742327
10 0.384500726990601
11 0.384160700686784
12 0.383775313437461
13 0.383331198444136
14 0.38281596042703
15 0.382218411685764
16 0.381528664583611
17 0.380738162434766
18 0.379839665650226
19 0.378827138654865
20 0.377695570463197
21 0.376440766693318
22 0.375059038846706
23 0.373546799653562
24 0.371900062661985
25 0.370113813332262
26 0.368181277062003
27 0.366093063306245
28 0.363836274716288
29 0.361393648645694
30 0.358742674026099
31 0.35585494100037
32 0.352695740441168
33 0.349223954249955
34 0.345392420048661
35 0.341148793933421
36 0.336437011272393
37 0.331199441652243
38 0.325379805921447
39 0.318927094584304
40 0.311800541326711
41 0.303975861991607
42 0.295452766849554
43 0.286263622227139
44 0.276482647280245
45 0.266234223974684
46 0.255697808429291
47 0.245105804452378
48 0.23473048492477
49 0.224858665029972
50 0.215759690406908
51 0.207660373310137
52 0.200740606158531
53 0.19514773692616
54 0.191010977615163
55 0.188443709034593
56 0.187539762029104
57 0.18837001553125
58 0.190979426255701
59 0.195384975408041
60 0.201573150055113
61 0.20949755430085
62 0.219075986754465
63 0.230187693389031
64 0.242672361053201
65 0.25633130702355
66 0.270927230913214
67 0.286187957969565
68 0.30181172740763
69 0.317474366211845
70 0.332838200734488
71 0.347563058985912
72 0.361318600439529
73 0.373797781257012
74 0.384729933715058
75 0.393892302992695
76 0.401118288936371
77 0.406301663744209
78 0.409396308597574
79 0.410411937255449
80 0.409407082775482
81 0.406481100764212
82 0.401766001384654
83 0.3954199539659
84 0.387621281574017
85 0.37856411095622
86 0.368452888941959
87 0.357498243625726
88 0.345909569040021
89 0.333889778395087
90 0.321628230622213
91 0.309296469559205
92 0.297044867750139
93 0.285001357370539
94 0.27327187245202
95 0.257764280910076
96 0.244542132822182
97 0.235440244069682
98 0.226651960075884
99 0.217583316626516
};

\nextgroupplot[
tick align=outside,
tick pos=left,
title={joint 1},
x grid style={darkgray176},
xmin=-4.95, xmax=103.95,
xtick style={color=black},
y grid style={darkgray176},
ymin=-0.185664212542998, ymax=0.357426361235277,
ytick style={color=black}
]
\addplot [semithick, blue]
table {%
0 0.174073845148087
1 0.173864498734474
2 0.17365662753582
3 0.173457831144333
4 0.173277750611305
5 0.173128247261047
6 0.173023447394371
7 0.172979831695557
8 0.173016041517258
9 0.173152685165405
10 0.173411965370178
11 0.17381726205349
12 0.174392536282539
13 0.175161734223366
14 0.176148071885109
15 0.177373185753822
16 0.178856283426285
17 0.180613115429878
18 0.182654827833176
19 0.184986665844917
20 0.187606573104858
21 0.190503731369972
22 0.193657070398331
23 0.197033956646919
24 0.200589090585709
25 0.204263895750046
26 0.207986548542976
27 0.211672559380531
28 0.215226233005524
29 0.2185427993536
30 0.221511155366898
31 0.224017024040222
32 0.225946232676506
33 0.227188169956207
34 0.227638900279999
35 0.227204114198685
36 0.225801900029182
37 0.223365128040314
38 0.219844073057175
39 0.215208858251572
40 0.209452107548714
41 0.202591836452484
42 0.194673955440521
43 0.185774818062782
44 0.176003023982048
45 0.16550001502037
46 0.154439270496368
47 0.143024206161499
48 0.131483405828476
49 0.120064944028854
50 0.109028868377209
51 0.0986389517784119
52 0.0891541987657547
53 0.0808210819959641
54 0.0738659724593163
55 0.0684898123145103
56 0.0648633167147636
57 0.0631237328052521
58 0.063372477889061
59 0.0656731277704239
60 0.0700499340891838
61 0.0764863565564156
62 0.0849234387278557
63 0.0952589064836502
64 0.107346720993519
65 0.120997622609138
66 0.135982021689415
67 0.152034193277359
68 0.168859407305717
69 0.186142385005951
70 0.203558146953583
71 0.220783084630966
72 0.237505957484245
73 0.253438621759415
74 0.268324375152588
75 0.281943917274475
76 0.294119775295258
77 0.304717302322388
78 0.313644826412201
79 0.320851564407349
80 0.326325505971909
81 0.330089807510376
82 0.332199990749359
83 0.332740426063538
84 0.331821233034134
85 0.329574763774872
86 0.326151967048645
87 0.321717798709869
88 0.316446781158447
89 0.310517549514771
90 0.304107248783112
91 0.297386348247528
92 0.290513277053833
93 0.283630609512329
94 0.276861608028412
95 0.27030873298645
96 0.26405268907547
97 0.258153468370438
98 0.252651393413544
99 0.247569680213928
};
\addplot [semithick, red]
table {%
0 0.181970879435539
1 0.19335150718689
2 0.207610681653023
3 0.213058575987816
4 0.215867757797241
5 0.218145236372948
6 0.221063479781151
7 0.224020272493362
8 0.227034837007523
9 0.230090424418449
10 0.233151540160179
11 0.236205607652664
12 0.239233791828156
13 0.242212757468224
14 0.245119750499725
15 0.247934356331825
16 0.250639647245407
17 0.25322163105011
18 0.255666106939316
19 0.257962942123413
20 0.260105013847351
21 0.262091845273972
22 0.263902425765991
23 0.265535324811935
24 0.266992300748825
25 0.268312901258469
26 0.269465863704681
27 0.2703837454319
28 0.271099507808685
29 0.271625548601151
30 0.271934807300568
31 0.272016048431396
32 0.271844327449799
33 0.27139675617218
34 0.270655989646912
35 0.269604325294495
36 0.268224567174911
37 0.266500681638718
38 0.264419466257095
39 0.26197350025177
40 0.25916188955307
41 0.255986303091049
42 0.252460479736328
43 0.248604878783226
44 0.244451522827148
45 0.240027382969856
46 0.23539873957634
47 0.230622157454491
48 0.225743293762207
49 0.220827654004097
50 0.215947329998016
51 0.211167201399803
52 0.206514313817024
53 0.20208166539669
54 0.197912305593491
55 0.194017931818962
56 0.190418973565102
57 0.187201619148254
58 0.184316635131836
59 0.181768968701363
60 0.17962321639061
61 0.177764713764191
62 0.176317751407623
63 0.175201058387756
64 0.174373939633369
65 0.173849523067474
66 0.173593416810036
67 0.173565551638603
68 0.173729449510574
69 0.174100756645203
70 0.174593269824982
71 0.175167500972748
72 0.175826832652092
73 0.176516577601433
74 0.177158504724503
75 0.177786082029343
76 0.178373530507088
77 0.178769528865814
78 0.179117411375046
79 0.179374113678932
80 0.179525926709175
81 0.179565474390984
82 0.179490566253662
83 0.179302975535393
84 0.179008021950722
85 0.17861345410347
86 0.178130611777306
87 0.177573591470718
88 0.176959410309792
89 0.17631421983242
90 0.175627544522285
91 0.174918100237846
92 0.174200564622879
93 0.173479676246643
94 0.172771602869034
95 0.172089636325836
96 0.171439036726952
97 0.17082367837429
98 0.170246213674545
99 0.170246213674545
};
\addplot [semithick, green01270]
table {%
0 0.176344663374573
1 0.176763131124373
2 0.17751139708666
3 0.178623390414709
4 0.180128458637505
5 0.18204923077592
6 0.184401615891659
7 0.187194327940364
8 0.190428447745145
9 0.194096963158669
10 0.19818437388045
11 0.202666441410287
12 0.207510156438022
13 0.212673976706305
14 0.21810838584126
15 0.223756774759869
16 0.229556614412493
17 0.235440877107661
18 0.241339622187097
19 0.247181668240454
20 0.252896240780838
21 0.258414506753805
22 0.263670912225211
23 0.268604245285217
24 0.273158378815421
25 0.277282664357
26 0.280931980503038
27 0.284066447582467
28 0.286650857407504
29 0.288653881938309
30 0.290047124693414
31 0.29080409678959
32 0.290899197749897
33 0.290306772303325
34 0.289000318010328
35 0.286951907229023
36 0.284131878898993
37 0.280508838256832
38 0.276050024263321
39 0.270722045441464
40 0.264492013989402
41 0.257329052328824
42 0.249206137658874
43 0.24010218456029
44 0.230004208948445
45 0.218909363882227
46 0.206826562025343
47 0.193777437612107
48 0.179796543047
49 0.164930892481471
50 0.149239272396844
51 0.132791793014184
52 0.115669963736263
53 0.0979671632266121
54 0.079789039856847
55 0.0612534301840835
56 0.0424896845559231
57 0.0236376113250431
58 0.00484620114141667
59 -0.0137278034993216
60 -0.0319214383693725
61 -0.0495671619430199
62 -0.0664946126666735
63 -0.0825326460990747
64 -0.0975111696391481
65 -0.111262966677406
66 -0.123626610192789
67 -0.134448853334724
68 -0.143587219338799
69 -0.150912687470565
70 -0.156312357140722
71 -0.159691981149725
72 -0.160978277371258
73 -0.160120918767573
74 -0.157094102547574
75 -0.151897598803718
76 -0.144557165178017
77 -0.135124243214526
78 -0.123674924594496
79 -0.110308216812717
80 -0.0951437061484141
81 -0.078318776251467
82 -0.0599855404371834
83 -0.0403076323703952
84 -0.0194570129146297
85 0.00238908992229172
86 0.0250510199248471
87 0.0483492378923961
88 0.0721065394178441
89 0.0961499468098576
90 0.120312280902604
91 0.144433388307111
92 0.168361308480311
93 0.191953002172789
94 0.215074862693404
95 0.220749416304154
96 0.2188448375664
97 0.216452073201579
98 0.213681352668626
99 0.210567293846658
};

\nextgroupplot[
tick align=outside,
tick pos=left,
title={joint 2},
x grid style={darkgray176},
xmin=-4.95, xmax=103.95,
xtick style={color=black},
y grid style={darkgray176},
ymin=-0.110655765250685, ymax=0.304307133456798,
ytick style={color=black}
]
\addplot [semithick, blue]
table {%
0 0.248909175395966
1 0.24870453774929
2 0.248455300927162
3 0.248152047395706
4 0.247783720493317
5 0.247337475419044
6 0.246798515319824
7 0.246150091290474
8 0.245373398065567
9 0.24444767832756
10 0.243350356817245
11 0.24205718934536
12 0.240542680025101
13 0.238780468702316
14 0.236743956804276
15 0.234407022595406
16 0.231745049357414
17 0.228736087679863
18 0.225362211465836
19 0.221611246466637
20 0.217478513717651
21 0.212968826293945
22 0.208098351955414
23 0.202896356582642
24 0.197406589984894
25 0.191688060760498
26 0.1858149766922
27 0.179875954985619
28 0.173972129821777
29 0.168214380741119
30 0.162719801068306
31 0.15760800242424
32 0.152996808290482
33 0.148998349905014
34 0.145715594291687
35 0.143239304423332
36 0.141645461320877
37 0.140993490815163
38 0.141324490308762
39 0.142659872770309
40 0.145000085234642
41 0.148323178291321
42 0.152583599090576
43 0.157711058855057
44 0.16360966861248
45 0.170158162713051
46 0.177210658788681
47 0.184598833322525
48 0.192135453224182
49 0.199618801474571
50 0.206838473677635
51 0.213581562042236
52 0.219638988375664
53 0.224811464548111
54 0.228915110230446
55 0.231785774230957
56 0.233282700181007
57 0.233291268348694
58 0.23172515630722
59 0.228528320789337
60 0.223676517605782
61 0.217179194092751
62 0.209080994129181
63 0.19946314394474
64 0.188444048166275
65 0.176179230213165
66 0.162858992815018
67 0.148705005645752
68 0.133964478969574
69 0.118902757763863
70 0.103794321417809
71 0.0889131873846054
72 0.0745233148336411
73 0.0608697235584259
74 0.0481709390878677
75 0.0366140305995941
76 0.0263504534959793
77 0.0174952000379562
78 0.0101266354322433
79 0.00428764522075653
80 -1.21742486953735e-05
81 -0.00279401242733002
82 -0.00410763919353485
83 -0.00402830541133881
84 -0.0026545375585556
85 -0.000104889273643494
86 0.0034850686788559
87 0.00796672701835632
88 0.0131825655698776
89 0.0189710855484009
90 0.0251718312501907
91 0.0316302925348282
92 0.0382028371095657
93 0.0447602421045303
94 0.0511908382177353
95 0.057402104139328
96 0.0633212774991989
97 0.0688947886228561
98 0.0740868896245956
99 0.078877717256546
};
\addplot [semithick, red]
table {%
0 0.250035136938095
1 0.249925434589386
2 0.249632105231285
3 0.253200381994247
4 0.251077115535736
5 0.25156831741333
6 0.250898241996765
7 0.250874310731888
8 0.250417828559875
9 0.250196397304535
10 0.249879121780396
11 0.249625831842422
12 0.249361246824265
13 0.249132484197617
14 0.248928561806679
15 0.248764321208
16 0.248644173145294
17 0.248577728867531
18 0.248566716909409
19 0.248617067933083
20 0.248733282089233
21 0.248914957046509
22 0.249166280031204
23 0.249474972486496
24 0.249845266342163
25 0.250302582979202
26 0.25084462761879
27 0.251420855522156
28 0.252077400684357
29 0.252807021141052
30 0.253604978322983
31 0.254467040300369
32 0.255400061607361
33 0.256404131650925
34 0.25748273730278
35 0.258638501167297
36 0.259875148534775
37 0.261196047067642
38 0.262604027986526
39 0.264099359512329
40 0.265681236982346
41 0.267346680164337
42 0.269086509943008
43 0.270889788866043
44 0.272736549377441
45 0.274599701166153
46 0.276465713977814
47 0.278268545866013
48 0.279999822378159
49 0.281579911708832
50 0.282970398664474
51 0.284103602170944
52 0.284882307052612
53 0.285304069519043
54 0.285445183515549
55 0.2850661277771
56 0.284313023090363
57 0.282707959413528
58 0.2806676030159
59 0.277925670146942
60 0.274580061435699
61 0.270610958337784
62 0.266008287668228
63 0.260890305042267
64 0.255159914493561
65 0.249034151434898
66 0.242425113916397
67 0.235625579953194
68 0.228570863604546
69 0.221449002623558
70 0.214441850781441
71 0.207542568445206
72 0.200934827327728
73 0.194710463285446
74 0.189177453517914
75 0.183971077203751
76 0.179546073079109
77 0.175791651010513
78 0.172874137759209
79 0.170739099383354
80 0.16943271458149
81 0.168945714831352
82 0.169280916452408
83 0.170413047075272
84 0.172302588820457
85 0.174893990159035
86 0.178120002150536
87 0.181899517774582
88 0.186143219470978
89 0.190780580043793
90 0.195674747228622
91 0.200780123472214
92 0.205986648797989
93 0.211225003004074
94 0.216418921947479
95 0.221508651971817
96 0.226442828774452
97 0.231182456016541
98 0.235696256160736
99 0.235696256160736
};
\addplot [semithick, green01270]
table {%
0 0.249885459603632
1 0.250121013011017
2 0.2506703202987
3 0.251386670147034
4 0.252204477809442
5 0.253072857270523
6 0.253960025116628
7 0.254846021598337
8 0.255719104825776
9 0.256572889083701
10 0.257404266887984
11 0.2582118603409
12 0.25899493848766
13 0.259752493677243
14 0.260482557113098
15 0.261181617782784
16 0.261844186237449
17 0.262462451474612
18 0.26302604177662
19 0.263521972829159
20 0.263934762315131
21 0.264246674397743
22 0.264438173657585
23 0.264488570679239
24 0.264376848146412
25 0.264082692168281
26 0.263587687421709
27 0.262876686987783
28 0.261939250844334
29 0.26077107100397
30 0.25937544737501
31 0.257764520731351
32 0.255960244750819
33 0.253995054040815
34 0.251912026307572
35 0.249764517589737
36 0.247615167383005
37 0.245534166304788
38 0.24359671939778
39 0.241879411560094
40 0.24045537452
41 0.239387929433985
42 0.238722546378504
43 0.238477039942379
44 0.238630426206196
45 0.239111798531725
46 0.239791951387291
47 0.24048212673032
48 0.240945087723626
49 0.240921214689526
50 0.2401645106536
51 0.23847327916651
52 0.23569854562518
53 0.231730052119498
54 0.226479661539982
55 0.219876766839269
56 0.211870519010451
57 0.202432163650831
58 0.191557047464863
59 0.179265746673627
60 0.165605832785819
61 0.150653724678345
62 0.134517346571992
63 0.117338416384981
64 0.0992934400471243
65 0.0805949380846533
66 0.0614911097821619
67 0.0422616389750302
68 0.0232121527368978
69 0.00466654843498475
70 -0.0130428933987173
71 -0.029587614490068
72 -0.044655941374459
73 -0.057966751527386
74 -0.0692819336364827
75 -0.0784165879392268
76 -0.0852454980855992
77 -0.0897055287026718
78 -0.0917938153094355
79 -0.0915624341718585
80 -0.089110882895251
81 -0.0845780540999755
82 -0.0781343804285243
83 -0.0699758385723957
84 -0.060318488458476
85 -0.0493945990845806
86 -0.0374474666758959
87 -0.0247273540112926
88 -0.0114838255222934
89 0.00203999453769428
90 0.0156152536519227
91 0.0290323416467486
92 0.0421053751052105
93 0.054674093962915
94 0.0666039555197165
95 0.081661826792516
96 0.094013319615642
97 0.101785810358928
98 0.108918871823764
99 0.116017351082368
};

\nextgroupplot[
tick align=outside,
tick pos=left,
title={joint 3},
x grid style={darkgray176},
xmin=-4.95, xmax=103.95,
xtick style={color=black},
y grid style={darkgray176},
ymin=-2.67090953736366, ymax=-1.91693408960012,
ytick style={color=black}
]
\addplot [semithick, blue]
table {%
0 -2.31558966636658
1 -2.30503940582275
2 -2.29356288909912
3 -2.28117704391479
4 -2.26792359352112
5 -2.2538697719574
6 -2.23910903930664
7 -2.22375988960266
8 -2.20796275138855
9 -2.19187355041504
10 -2.17565941810608
11 -2.15948939323425
12 -2.14352774620056
13 -2.12792658805847
14 -2.11282086372375
15 -2.09832286834717
16 -2.08452129364014
17 -2.07148098945618
18 -2.05924415588379
19 -2.04783463478088
20 -2.03726077079773
21 -2.02752232551575
22 -2.01861381530762
23 -2.01053071022034
24 -2.00327324867249
25 -1.99685049057007
26 -1.99128293991089
27 -1.98660385608673
28 -1.98286104202271
29 -1.98011553287506
30 -1.97844076156616
31 -1.97792136669159
32 -1.97864973545074
33 -1.98072421550751
34 -1.98424553871155
35 -1.98931348323822
36 -1.99602341651917
37 -2.00446176528931
38 -2.01470232009888
39 -2.02679991722107
40 -2.04078650474548
41 -2.05666351318359
42 -2.07439851760864
43 -2.09391808509827
44 -2.11510562896729
45 -2.13779902458191
46 -2.16179132461548
47 -2.18683505058289
48 -2.21264934539795
49 -2.238929271698
50 -2.2653591632843
51 -2.29162430763245
52 -2.3174262046814
53 -2.34249186515808
54 -2.36658620834351
55 -2.38951778411865
56 -2.41114091873169
57 -2.43135809898376
58 -2.45011568069458
59 -2.46740007400513
60 -2.48323225975037
61 -2.49765920639038
62 -2.51074981689453
63 -2.52258658409119
64 -2.53326034545898
65 -2.54286599159241
66 -2.55149865150452
67 -2.559250831604
68 -2.56620955467224
69 -2.57245707511902
70 -2.57806873321533
71 -2.58311367034912
72 -2.58765459060669
73 -2.5917489528656
74 -2.59544849395752
75 -2.59880018234253
76 -2.60184764862061
77 -2.60462975502014
78 -2.60718178749084
79 -2.60953593254089
80 -2.61172080039978
81 -2.61376094818115
82 -2.61567831039429
83 -2.6174910068512
84 -2.61921310424805
85 -2.62085580825806
86 -2.62242722511292
87 -2.62393188476562
88 -2.62537264823914
89 -2.62674927711487
90 -2.62806057929993
91 -2.62930488586426
92 -2.63047933578491
93 -2.63158178329468
94 -2.63260984420776
95 -2.63356328010559
96 -2.63444137573242
97 -2.63524484634399
98 -2.63597631454468
99 -2.63663792610168
};
\addplot [semithick, red]
table {%
0 -2.39721632003784
1 -2.3943202495575
2 -2.39192199707031
3 -2.38008189201355
4 -2.37473273277283
5 -2.37033772468567
6 -2.3646674156189
7 -2.35903835296631
8 -2.35329532623291
9 -2.34753084182739
10 -2.3417866230011
11 -2.33609414100647
12 -2.33049130439758
13 -2.3250150680542
14 -2.31970596313477
15 -2.31459736824036
16 -2.30971431732178
17 -2.30507802963257
18 -2.30071020126343
19 -2.29662394523621
20 -2.29283165931702
21 -2.2893328666687
22 -2.28614616394043
23 -2.28328227996826
24 -2.28073191642761
25 -2.27844524383545
26 -2.27646374702454
27 -2.27489519119263
28 -2.2736918926239
29 -2.27283716201782
30 -2.2723696231842
31 -2.27231454849243
32 -2.27271699905396
33 -2.27361726760864
34 -2.27504801750183
35 -2.27704453468323
36 -2.27964019775391
37 -2.28286981582642
38 -2.2867591381073
39 -2.29132771492004
40 -2.29658889770508
41 -2.30254316329956
42 -2.3091778755188
43 -2.31646418571472
44 -2.32435727119446
45 -2.33278632164001
46 -2.34166121482849
47 -2.35090684890747
48 -2.36041808128357
49 -2.37006711959839
50 -2.37972736358643
51 -2.38926911354065
52 -2.39854288101196
53 -2.40745186805725
54 -2.41590738296509
55 -2.42376565933228
56 -2.43102264404297
57 -2.43750143051147
58 -2.4432213306427
59 -2.44815516471863
60 -2.45230913162231
61 -2.45564961433411
62 -2.45819568634033
63 -2.46001243591309
64 -2.46115398406982
65 -2.46166682243347
66 -2.46160459518433
67 -2.46105766296387
68 -2.46005773544312
69 -2.45871448516846
70 -2.45710229873657
71 -2.45530271530151
72 -2.45341396331787
73 -2.45150399208069
74 -2.4497241973877
75 -2.44802808761597
76 -2.44647979736328
77 -2.4452428817749
78 -2.44428682327271
79 -2.44358396530151
80 -2.44316411018372
81 -2.44304013252258
82 -2.44321751594543
83 -2.44369006156921
84 -2.44444513320923
85 -2.44546270370483
86 -2.44671678543091
87 -2.44817566871643
88 -2.44979929924011
89 -2.45149922370911
90 -2.45328259468079
91 -2.45514273643494
92 -2.45704174041748
93 -2.45894956588745
94 -2.46084094047546
95 -2.46267771720886
96 -2.46445274353027
97 -2.46614909172058
98 -2.46775794029236
99 -2.46775794029236
};
\addplot [semithick, green01270]
table {%
0 -2.39832045814955
1 -2.39595188762615
2 -2.39196665698978
3 -2.38632027360233
4 -2.37899252959041
5 -2.36998723701364
6 -2.35933167970682
7 -2.34707627777847
8 -2.33329423932595
9 -2.31808109136097
10 -2.30155392961304
11 -2.28385020969183
12 -2.26512601441089
13 -2.24555378152157
14 -2.22531947373457
15 -2.20461928207651
16 -2.1836560248703
17 -2.16263534119636
18 -2.14176186592451
19 -2.12123551613665
20 -2.10124807814434
21 -2.08198019366656
22 -2.06359885773905
23 -2.04625553807065
24 -2.03008494289614
25 -2.0152044757778
26 -2.00171434661329
27 -1.9896983232372
28 -1.97922502228289
29 -1.97034963107718
30 -1.96311599025712
31 -1.9575588548918
32 -1.95370620288514
33 -1.95158149337331
34 -1.9512057008621
35 -1.95259899117408
36 -1.95578190904583
37 -1.96077597589831
38 -1.96760348185296
39 -1.97628639718471
40 -1.98684424424573
41 -1.99929081878382
42 -2.0136297092406
43 -2.0298487258934
44 -2.0479135766696
45 -2.06776145701901
46 -2.08929580784081
47 -2.11238378457854
48 -2.13685788440363
49 -2.16252232851092
50 -2.1891626335982
51 -2.21655487274213
52 -2.24447101873258
53 -2.27267973080706
54 -2.30094552713181
55 -2.32902938346726
56 -2.35669120071803
57 -2.38369307721037
58 -2.40980267482609
59 -2.43479649515278
60 -2.45846302154287
61 -2.48060577037587
62 -2.5010462561107
63 -2.51962756770197
64 -2.53621610156276
65 -2.55070289917786
66 -2.5630088067206
67 -2.57308573371788
68 -2.58091802539941
69 -2.58652337038912
70 -2.58995279582124
71 -2.59128970831631
72 -2.59064802121628
73 -2.58816949590769
74 -2.58402056840331
75 -2.57838888626123
76 -2.57147971714093
77 -2.56351232590554
78 -2.55471641872421
79 -2.54532859858054
80 -2.53558877915312
81 -2.52573655298487
82 -2.51600747459576
83 -2.50662925325902
84 -2.49781796989572
85 -2.48977438011849
86 -2.48268042086095
87 -2.47669606669336
88 -2.47195663365504
89 -2.46857072230334
90 -2.46661894093485
91 -2.46615355304796
92 -2.46719869501136
93 -2.46975179765609
94 -2.47378574899013
95 -2.47265599806276
96 -2.4711930303528
97 -2.47190352081367
98 -2.4745025788828
99 -2.47880306349562
};

\nextgroupplot[
tick align=outside,
tick pos=left,
title={joint 4},
x grid style={darkgray176},
xmin=-4.95, xmax=103.95,
xtick style={color=black},
y grid style={darkgray176},
ymin=-0.23445477364195, ymax=0.0997367818385241,
ytick style={color=black}
]
\addplot [semithick, blue]
table {%
0 -0.0273089520633221
1 -0.0209375619888306
2 -0.0140944123268127
3 -0.00682199001312256
4 0.000815697014331818
5 0.00873130559921265
6 0.0168137848377228
7 0.0249286964535713
8 0.0329207256436348
9 0.0406169891357422
10 0.0478318631649017
11 0.0543724894523621
12 0.0600451678037643
13 0.0646613389253616
14 0.0680433958768845
15 0.0700301080942154
16 0.0704809129238129
17 0.0692799985408783
18 0.0663397312164307
19 0.0616038143634796
20 0.0550503581762314
21 0.0466948747634888
22 0.0365931689739227
23 0.0248436480760574
24 0.0115889087319374
25 -0.00298407673835754
26 -0.0186459869146347
27 -0.0351289547979832
28 -0.0521330013871193
29 -0.0693349465727806
30 -0.0863989070057869
31 -0.102987378835678
32 -0.118773117661476
33 -0.1334498077631
34 -0.146740838885307
35 -0.158406749367714
36 -0.168249905109406
37 -0.176117137074471
38 -0.18190099298954
39 -0.185539796948433
40 -0.187016755342484
41 -0.186359122395515
42 -0.183636963367462
43 -0.178962111473083
44 -0.172486484050751
45 -0.164400011301041
46 -0.154927045106888
47 -0.144322037696838
48 -0.132862776517868
49 -0.120843052864075
50 -0.108563534915447
51 -0.0963223651051521
52 -0.0844058096408844
53 -0.0730801597237587
54 -0.0625841170549393
55 -0.0531240440905094
56 -0.0448704175651073
57 -0.0379564389586449
58 -0.0324778854846954
59 -0.0284941308200359
60 -0.0260296314954758
61 -0.0250758528709412
62 -0.0255931802093983
63 -0.0275128148496151
64 -0.0307388156652451
65 -0.0351503454148769
66 -0.0406045727431774
67 -0.0469402857124805
68 -0.0539822578430176
69 -0.0615464597940445
70 -0.069445937871933
71 -0.0774968340992928
72 -0.0855241343379021
73 -0.0933670923113823
74 -0.100883357226849
75 -0.107951804995537
76 -0.114474415779114
77 -0.120376423001289
78 -0.125605672597885
79 -0.130131274461746
80 -0.133941531181335
81 -0.137041762471199
82 -0.139452069997787
83 -0.141205072402954
84 -0.142343968153
85 -0.142920404672623
86 -0.142992734909058
87 -0.142624005675316
88 -0.141880080103874
89 -0.140827670693398
90 -0.139532446861267
91 -0.138057246804237
92 -0.13646037876606
93 -0.134794414043427
94 -0.133105158805847
95 -0.131431207060814
96 -0.129803746938705
97 -0.128246784210205
98 -0.126777738332748
99 -0.125408127903938
};
\addplot [semithick, red]
table {%
0 -0.0765574425458908
1 -0.0737884044647217
2 -0.0708136782050133
3 -0.0685353130102158
4 -0.0628875121474266
5 -0.060865331441164
6 -0.0579647049307823
7 -0.0549074821174145
8 -0.0519197881221771
9 -0.0490739084780216
10 -0.0462262332439423
11 -0.043533269315958
12 -0.0410137809813023
13 -0.038718618452549
14 -0.0366658456623554
15 -0.0348830036818981
16 -0.0333954878151417
17 -0.032241128385067
18 -0.03142961114645
19 -0.0309588965028524
20 -0.0308268666267395
21 -0.0310347154736519
22 -0.031564611941576
23 -0.0324446819722652
24 -0.0336830541491508
25 -0.0353037007153034
26 -0.0370955243706703
27 -0.0390129946172237
28 -0.0410326607525349
29 -0.0432139337062836
30 -0.0454849079251289
31 -0.0477546863257885
32 -0.0499532297253609
33 -0.052024882286787
34 -0.0539265461266041
35 -0.0556198731064796
36 -0.0570742152631283
37 -0.0582659766077995
38 -0.0591795481741428
39 -0.0598035790026188
40 -0.0601351000368595
41 -0.0601764693856239
42 -0.059936948120594
43 -0.0594323612749577
44 -0.0586854591965675
45 -0.0577528178691864
46 -0.0566907152533531
47 -0.0555305629968643
48 -0.0542523451149464
49 -0.0529254600405693
50 -0.0516011342406273
51 -0.0503414943814278
52 -0.0494610145688057
53 -0.0487786903977394
54 -0.0479740686714649
55 -0.0473399721086025
56 -0.046766109764576
57 -0.0468440055847168
58 -0.0472658313810825
59 -0.0479103811085224
60 -0.0488055720925331
61 -0.0498301796615124
62 -0.0509930215775967
63 -0.0524388290941715
64 -0.0542161129415035
65 -0.056223139166832
66 -0.0585416667163372
67 -0.0609764456748962
68 -0.0636227428913116
69 -0.0661445483565331
70 -0.0684173479676247
71 -0.070818617939949
72 -0.0730468407273293
73 -0.0750400647521019
74 -0.0770446136593819
75 -0.0789141207933426
76 -0.0805360376834869
77 -0.0818145647644997
78 -0.0827028676867485
79 -0.0833964794874191
80 -0.0838988944888115
81 -0.0841736346483231
82 -0.0842254832386971
83 -0.0840719193220139
84 -0.0837421342730522
85 -0.0832582339644432
86 -0.0826433077454567
87 -0.0819229185581207
88 -0.0811410695314407
89 -0.0805178135633469
90 -0.0798966959118843
91 -0.0792196616530418
92 -0.0785110741853714
93 -0.0777843147516251
94 -0.0770393460988998
95 -0.0763365626335144
96 -0.0756426602602005
97 -0.074965737760067
98 -0.0743177533149719
99 -0.0743177533149719
};
\addplot [semithick, green01270]
table {%
0 -0.0780481824494494
1 -0.0734071201595968
2 -0.0666516283653648
3 -0.0582995785690491
4 -0.0487844761841865
5 -0.0384544984657076
6 -0.0275932725322714
7 -0.0164352962898236
8 -0.00517942487871942
9 0.00600067592244222
10 0.0169488106614567
11 0.0275207628305885
12 0.0375804596113939
13 0.0469973732298164
14 0.055644732656838
15 0.0633983264212946
16 0.0701358157431154
17 0.0757365183345495
18 0.0800817097814599
19 0.0830554840554701
20 0.0845462565894116
21 0.0844489699926198
22 0.0826679930397801
23 0.079120684644934
24 0.0737415344722298
25 0.0664867080738962
26 0.057338820379259
27 0.046311701883346
28 0.0334549248844307
29 0.0188578683164278
30 0.0026530772808543
31 -0.0149811774105169
32 -0.0338193224131347
33 -0.0535880149444376
34 -0.0739669067037642
35 -0.0945908804557428
36 -0.115054068900026
37 -0.134916090901907
38 -0.153711242983942
39 -0.170961356180231
40 -0.186193338024294
41 -0.198962249077592
42 -0.208880469240137
43 -0.215652513633988
44 -0.219113240262461
45 -0.219264248392838
46 -0.216299294651275
47 -0.210606006940277
48 -0.202731255995944
49 -0.193305343749413
50 -0.182939472017458
51 -0.172133320683588
52 -0.161234327422852
53 -0.150459896684556
54 -0.139949704598519
55 -0.129807031245208
56 -0.120118378370499
57 -0.11096155176186
58 -0.102409923867767
59 -0.0945349154756023
60 -0.0874064161756672
61 -0.0810912916276198
62 -0.0756498987727326
63 -0.0711262426245544
64 -0.067552168504222
65 -0.0649438489980584
66 -0.0632752234042176
67 -0.0624837323968748
68 -0.0624714190348027
69 -0.0631075117570453
70 -0.0642365069563707
71 -0.0656906415438034
72 -0.0673046312007639
73 -0.0689300348710926
74 -0.0704463315447045
75 -0.0717673124850915
76 -0.0728422140571026
77 -0.0736525175917796
78 -0.0742055138532226
79 -0.0745264342021581
80 -0.0746506596579918
81 -0.0746169721138703
82 -0.0744623688072801
83 -0.0742186758309526
84 -0.0739105736097099
85 -0.0735549471523426
86 -0.0731613244706212
87 -0.0727332574361733
88 -0.0722707609429571
89 -0.0717733604838471
90 -0.0712438611172115
91 -0.0706920084890877
92 -0.0701371647178285
93 -0.0696099775348928
94 -0.069152085535625
95 -0.0752364845929344
96 -0.080065046056118
97 -0.0814173571476026
98 -0.0817804335162243
99 -0.0818318375204557
};

\nextgroupplot[
tick align=outside,
tick pos=left,
title={joint 5},
x grid style={darkgray176},
xmin=-4.95, xmax=103.95,
xtick style={color=black},
y grid style={darkgray176},
ymin=1.99667933505055, ymax=3.3413681156549,
ytick style={color=black}
]
\addplot [semithick, blue]
table {%
0 2.47220373153687
1 2.46025085449219
2 2.44732022285461
3 2.4334568977356
4 2.41874027252197
5 2.40328431129456
6 2.3872401714325
7 2.37079334259033
8 2.35416102409363
9 2.33758544921875
10 2.3213255405426
11 2.3056492805481
12 2.29082202911377
13 2.27709865570068
14 2.26471519470215
15 2.25388169288635
16 2.24477767944336
17 2.23755025863647
18 2.23231196403503
19 2.2291419506073
20 2.22808599472046
21 2.22915935516357
22 2.23234724998474
23 2.23760771751404
24 2.24487400054932
25 2.25405693054199
26 2.26504921913147
27 2.27772974967957
28 2.29196929931641
29 2.30763673782349
30 2.32460641860962
31 2.34276533126831
32 2.36201906204224
33 2.38229608535767
34 2.40355181694031
35 2.42576813697815
36 2.44895076751709
37 2.47312378883362
38 2.49832344055176
39 2.52458524703979
40 2.55193495750427
41 2.5803747177124
42 2.60987067222595
43 2.64034175872803
44 2.67165040969849
45 2.70359683036804
46 2.73591542243958
47 2.76827883720398
48 2.80030417442322
49 2.83156418800354
50 2.86160326004028
51 2.88995385169983
52 2.91615414619446
53 2.93976521492004
54 2.96038675308228
55 2.97766757011414
56 2.99131631851196
57 3.00110578536987
58 3.0068793296814
59 3.0085506439209
60 3.00610733032227
61 2.99961113929749
62 2.9891996383667
63 2.97508668899536
64 2.95756053924561
65 2.93698453903198
66 2.91378903388977
67 2.88846468925476
68 2.86154985427856
69 2.83361554145813
70 2.80524730682373
71 2.77702498435974
72 2.74950575828552
73 2.72320437431335
74 2.69857931137085
75 2.67602324485779
76 2.65585517883301
77 2.63831901550293
78 2.62358331680298
79 2.61174511909485
80 2.60283350944519
81 2.59681630134583
82 2.59360456466675
83 2.59305858612061
84 2.59499359130859
85 2.59918594360352
86 2.6053786277771
87 2.61329078674316
88 2.62262392044067
89 2.63307189941406
90 2.64433073997498
91 2.65610861778259
92 2.66813254356384
93 2.68015837669373
94 2.69197416305542
95 2.70340394973755
96 2.71430945396423
97 2.72458791732788
98 2.73417091369629
99 2.74301886558533
};
\addplot [semithick, red]
table {%
0 2.56267356872559
1 2.5580780506134
2 2.55314350128174
3 2.53245806694031
4 2.52584838867188
5 2.52001476287842
6 2.5125675201416
7 2.50519824028015
8 2.49774312973022
9 2.49025630950928
10 2.48283076286316
11 2.47550129890442
12 2.4683256149292
13 2.46134901046753
14 2.45461273193359
15 2.44816136360168
16 2.44202041625977
17 2.43620920181274
18 2.4307496547699
19 2.42565989494324
20 2.4209463596344
21 2.41663527488708
22 2.41269540786743
23 2.40916419029236
24 2.40601348876953
25 2.40327787399292
26 2.40089917182922
27 2.39901566505432
28 2.39758729934692
29 2.39662742614746
30 2.39614868164062
31 2.39618754386902
32 2.39679336547852
33 2.39801955223083
34 2.39990663528442
35 2.4025022983551
36 2.40585160255432
37 2.41000270843506
38 2.41499471664429
39 2.4208562374115
40 2.42761445045471
41 2.43527626991272
42 2.44383692741394
43 2.45326948165894
44 2.46353030204773
45 2.47452163696289
46 2.48614382743835
47 2.49833583831787
48 2.51094079017639
49 2.52380084991455
50 2.53675413131714
51 2.54963397979736
52 2.56223106384277
53 2.57440638542175
54 2.58603978157043
55 2.59690046310425
56 2.60697293281555
57 2.61601996421814
58 2.62401342391968
59 2.63091421127319
60 2.63677310943604
61 2.64142417907715
62 2.64500832557678
63 2.64753913879395
64 2.64910864830017
65 2.64976668357849
66 2.64963150024414
67 2.64876961708069
68 2.64727258682251
69 2.64533734321594
70 2.64297938346863
71 2.64035105705261
72 2.63763213157654
73 2.6348922252655
74 2.63228821754456
75 2.62983250617981
76 2.6275851726532
77 2.62573623657227
78 2.62435150146484
79 2.62333226203918
80 2.62271666526794
81 2.62253165245056
82 2.6227810382843
83 2.62345623970032
84 2.62453436851501
85 2.62598824501038
86 2.62777829170227
87 2.62986135482788
88 2.63218092918396
89 2.63459086418152
90 2.63710331916809
91 2.63973021507263
92 2.64241528511047
93 2.64510774612427
94 2.64778065681458
95 2.65037822723389
96 2.65289187431335
97 2.65530061721802
98 2.65758466720581
99 2.65758466720581
};
\addplot [semithick, green01270]
table {%
0 2.56665630498147
1 2.56125137840524
2 2.55240343030801
3 2.54021709492626
4 2.52484101364873
5 2.50646837149999
6 2.4853349239895
7 2.46171752632031
8 2.43593214963883
9 2.40833111825161
10 2.37929938344802
11 2.34924957885454
12 2.31861591762808
13 2.28784715323862
14 2.25739875855235
15 2.22772469374214
16 2.19926923221112
17 2.17245909782153
18 2.14769625984115
19 2.12535156906697
20 2.10575945399508
21 2.08921368596743
22 2.07596423613556
23 2.06621527061937
24 2.06012416799327
25 2.05780155235074
26 2.05931226761253
27 2.06467725983991
28 2.07387626203479
29 2.08685120719941
30 2.10351033787962
31 2.12373276976628
32 2.1473733096401
33 2.17426742434255
34 2.20423603004867
35 2.23708980094866
36 2.27263277470064
37 2.3106650992894
38 2.35098453719818
39 2.39338674125915
40 2.43766424453335
41 2.48360410552959
42 2.5309843735397
43 2.5795696370453
44 2.62910596104896
45 2.67931559812149
46 2.7298923767792
47 2.78049876960634
48 2.83076586785886
49 2.88029759962542
50 2.92867909992307
51 2.97548734235077
52 3.02030085561884
53 3.06270639445367
54 3.102303587224
55 3.13871016182971
56 3.17156849313498
57 3.20055251678275
58 3.22537406096827
59 3.24578836921339
60 3.26159885802543
61 3.27266120936531
62 3.27888692052686
63 3.2802458983547
64 3.27676886941041
65 3.2685508880779
66 3.25575159349708
67 3.23859491850978
68 3.21736800049078
69 3.19241868416657
70 3.16415117486959
71 3.13301995180859
72 3.09952202102319
73 3.06418769768546
74 3.02757043749502
75 2.99023606572908
76 2.95275180311376
77 2.91567541937911
78 2.87954489448173
79 2.84486873663385
80 2.81211711512079
81 2.78171392784406
82 2.75402979287245
83 2.72937601022919
84 2.70799960047686
85 2.6900794266704
86 2.67572360223013
87 2.66496829581498
88 2.65777799052038
89 2.65404748031221
90 2.65360563811601
91 2.65622056456125
92 2.66160729543377
93 2.66943615734656
94 2.67934235798983
95 2.67688963208081
96 2.67178091089407
97 2.67089028123034
98 2.67351702658322
99 2.67920435251635
};

\nextgroupplot[
tick align=outside,
tick pos=left,
title={joint 6},
x grid style={darkgray176},
xmin=-4.95, xmax=103.95,
xtick style={color=black},
y grid style={darkgray176},
ymin=1.19632429623383, ymax=1.72570630850455,
ytick style={color=black}
]
\addplot [semithick, blue]
table {%
0 1.43312156200409
1 1.42732071876526
2 1.42102956771851
3 1.41426455974579
4 1.40705692768097
5 1.39945375919342
6 1.39151835441589
7 1.38332974910736
8 1.37498021125793
9 1.36657297611237
10 1.35821831226349
11 1.35002911090851
12 1.34211659431458
13 1.33458614349365
14 1.32753312587738
15 1.32104063034058
16 1.31517767906189
17 1.30999791622162
18 1.30554032325745
19 1.30182945728302
20 1.29887700080872
21 1.29668343067169
22 1.29523944854736
23 1.29452776908875
24 1.29452478885651
25 1.29520189762115
26 1.29652774333954
27 1.29846906661987
28 1.30099308490753
29 1.30406880378723
30 1.30766844749451
31 1.31176996231079
32 1.31635713577271
33 1.32142114639282
34 1.32696044445038
35 1.33298075199127
36 1.3394935131073
37 1.34651446342468
38 1.35406029224396
39 1.36214661598206
40 1.37078297138214
41 1.37997031211853
42 1.38969612121582
43 1.39993155002594
44 1.41062819957733
45 1.42171669006348
46 1.43310618400574
47 1.44468533992767
48 1.45632517337799
49 1.46788346767426
50 1.47920942306519
51 1.49015045166016
52 1.50055778026581
53 1.51029229164124
54 1.51922965049744
55 1.52726376056671
56 1.53430914878845
57 1.54030251502991
58 1.54520285129547
59 1.54899096488953
60 1.55166900157928
61 1.55325853824615
62 1.55380058288574
63 1.55335319042206
64 1.55199134349823
65 1.54980492591858
66 1.54689693450928
67 1.54338181018829
68 1.53938210010529
69 1.53502559661865
70 1.53044104576111
71 1.52575540542603
72 1.52108919620514
73 1.51655352115631
74 1.51224720478058
75 1.5082551240921
76 1.50464689731598
77 1.50147640705109
78 1.49878263473511
79 1.49658966064453
80 1.49490821361542
81 1.49373662471771
82 1.49306213855743
83 1.49286222457886
84 1.49310553073883
85 1.49375355243683
86 1.4947612285614
87 1.49607908725739
88 1.49765419960022
89 1.49943256378174
90 1.50136017799377
91 1.50338506698608
92 1.50545871257782
93 1.5075376033783
94 1.50958395004272
95 1.51156651973724
96 1.51346027851105
97 1.51524686813354
98 1.51691389083862
99 1.51845407485962
};
\addplot [semithick, red]
table {%
0 1.47823619842529
1 1.47714555263519
2 1.47601425647736
3 1.47181236743927
4 1.47039294242859
5 1.46946394443512
6 1.46788966655731
7 1.46668541431427
8 1.46557319164276
9 1.46445047855377
10 1.46338295936584
11 1.46238839626312
12 1.4614759683609
13 1.46064674854279
14 1.45989751815796
15 1.45921313762665
16 1.45860636234283
17 1.45809578895569
18 1.4576587677002
19 1.45728433132172
20 1.45696783065796
21 1.45670342445374
22 1.45652377605438
23 1.45650207996368
24 1.45657134056091
25 1.45654952526093
26 1.45651030540466
27 1.45672559738159
28 1.45706927776337
29 1.4571281671524
30 1.45724081993103
31 1.45738279819489
32 1.45753943920135
33 1.45770704746246
34 1.45790076255798
35 1.45812559127808
36 1.45838928222656
37 1.45869028568268
38 1.45903551578522
39 1.45943975448608
40 1.45990419387817
41 1.46042847633362
42 1.46101605892181
43 1.46166062355042
44 1.46236002445221
45 1.46315264701843
46 1.4640154838562
47 1.46495759487152
48 1.46585190296173
49 1.4667694568634
50 1.46768033504486
51 1.46861326694489
52 1.46957969665527
53 1.47017920017242
54 1.47094202041626
55 1.47203660011292
56 1.47295248508453
57 1.4739431142807
58 1.47488927841187
59 1.47583603858948
60 1.47690200805664
61 1.47815144062042
62 1.47955501079559
63 1.4805144071579
64 1.48177933692932
65 1.48243021965027
66 1.48361873626709
67 1.48410201072693
68 1.48492407798767
69 1.48599100112915
70 1.48626124858856
71 1.48649454116821
72 1.48720836639404
73 1.48821628093719
74 1.4875351190567
75 1.48825931549072
76 1.48837769031525
77 1.48856091499329
78 1.48880302906036
79 1.48902499675751
80 1.48930597305298
81 1.48963630199432
82 1.48995316028595
83 1.49018013477325
84 1.49027061462402
85 1.49019587039948
86 1.48994541168213
87 1.48951935768127
88 1.48893868923187
89 1.48832130432129
90 1.48762667179108
91 1.48684680461884
92 1.48598897457123
93 1.48506808280945
94 1.48409283161163
95 1.48309755325317
96 1.48208725452423
97 1.48107469081879
98 1.48007166385651
99 1.48007166385651
};
\addplot [semithick, green01270]
table {%
0 1.4777938807129
1 1.47341676333397
2 1.46707093545238
3 1.45925107038713
4 1.4503639027946
5 1.44072879163001
6 1.43059892027174
7 1.42017673998527
8 1.40962707881602
9 1.39908693375678
10 1.38867232652681
11 1.37848283685345
12 1.36860448856991
13 1.35911149895751
14 1.35006744218112
15 1.34152615789211
16 1.33353257488336
17 1.32612359627847
18 1.31932906529352
19 1.31317281119249
20 1.30767371455608
21 1.30284668246317
22 1.29870345396059
23 1.2952531189089
24 1.29250222397669
25 1.29045439440183
26 1.2891093529129
27 1.28846129387926
28 1.28849659102746
29 1.28919084884315
30 1.29050547156286
31 1.29238381621628
32 1.2947471943531
33 1.29749104193902
34 1.30048160134499
35 1.3035535986137
36 1.30650949454096
37 1.30912097447074
38 1.31113362654413
39 1.31227570598128
40 1.31227216985968
41 1.31086496193271
42 1.30784017929984
43 1.30306170270251
44 1.29650888759468
45 1.28831276693528
46 1.27878099645695
47 1.26839806502135
48 1.2577875290732
49 1.24763155360208
50 1.2385635688079
51 1.23107308171659
52 1.22546570161471
53 1.22188872012176
54 1.22038711497341
55 1.22094793894429
56 1.22352301973061
57 1.22804061268994
58 1.23441365559399
59 1.24254649791298
60 1.25233965863543
61 1.26369264840375
62 1.27650463420355
63 1.29066812963881
64 1.30607634603246
65 1.3226229350058
66 1.34017610021201
67 1.35858402348942
68 1.37767452908176
69 1.3972551198524
70 1.41711765820246
71 1.43704707535374
72 1.45683260823804
73 1.47627942355648
74 1.49521810695961
75 1.51351069405105
76 1.53105254553259
77 1.54777065640185
78 1.56361907819893
79 1.57857285113211
80 1.59262164343687
81 1.60576388373308
82 1.61800191045112
83 1.6293384270483
84 1.63977407989
85 1.64930619042854
86 1.65792858418113
87 1.66563243618966
88 1.67240833591925
89 1.67824909684124
90 1.68315342196154
91 1.68712943084533
92 1.69019773042564
93 1.69239309971955
94 1.69376432657713
95 1.6989250835009
96 1.70164348976497
97 1.7003212237092
98 1.69740414578392
99 1.69357362314658
};
\end{groupplot}

\end{tikzpicture}

%% file: tikz/legend_act_traj.tex
\begin{tikzpicture} 
    \begin{axis}[%
    hide axis,
    xmin=10,
    xmax=50,
    ymin=0,
    ymax=0.4,
    legend style={
        draw=white!15!black,
        legend cell align=left,
        legend columns=-1, 
        legend style={
            draw=none,
            column sep=1ex,
            line width=1pt
        }
    },
    ]
    \addlegendimage{blue}
    \addlegendentry{Reference trajectory};
    \addlegendimage{C4}
    \addlegendentry{Actual trajectory: Fancy Gym};
    \addlegendimage{red}
    \addlegendentry{Actual trajectory: Orbit};
    \end{axis}
\end{tikzpicture}

%% file: tikz/box_pushing_FG_HP.tex
% This file was created with tikzplotlib v0.10.1.
\begin{tikzpicture}

\begin{axis}[
legend cell align={left},
legend style={fill opacity=0.8, draw opacity=1, text opacity=1, draw=lightgray204, at={(0.03,0.03)},  anchor=north west},
title={Box Pushing - Dense reward},
% every x tick scale label/.style={
%     at={(1,0)},yshift=5pt, xshift=-5pt,anchor=south east,inner sep=0pt
% },
tick align=outside,
tick pos=left,
x grid style={darkgray176},
xlabel={Number Environment Interactions},
xmajorgrids,
xmin=-2500000, xmax=42000000.05,
xtick style={color=black},
y grid style={darkgray176},
ylabel={Success Rate},
ymajorgrids,
ymin=-0.05, ymax=1.05,
ytick style={color=black}
]

\path [draw=C4, fill=C4, opacity=0.2]
(axis cs:32000,0)
--(axis cs:32000,0)
--(axis cs:1408000,0.0635725795894805)
--(axis cs:2784000,0.250919000510674)
--(axis cs:4160000,0.470246656893603)
--(axis cs:5536000,0.592102295659112)
--(axis cs:6912000,0.782791550540178)
--(axis cs:8288000,0.872748789297704)
--(axis cs:9664000,0.847595560835828)
--(axis cs:11040000,0.886797518739585)
--(axis cs:12416000,0.912747751408863)
--(axis cs:13808000,0.918796697684473)
--(axis cs:15184000,0.901289934131986)
--(axis cs:16560000,0.925059064504919)
--(axis cs:17936000,0.917931024831256)
--(axis cs:19312000,0.914597308239019)
--(axis cs:20688000,0.888485146435793)
--(axis cs:22064000,0.90531941886275)
--(axis cs:23440000,0.919090924957461)
--(axis cs:24816000,0.898246249155049)
--(axis cs:26192000,0.881525980154551)
--(axis cs:27584000,0.889175468643881)
--(axis cs:28960000,0.867545652077287)
--(axis cs:30336000,0.853060339771584)
--(axis cs:31712000,0.772822677081848)
--(axis cs:33088000,0.840994058053027)
--(axis cs:34464000,0.80094362154582)
--(axis cs:35840000,0.789689147852871)
--(axis cs:37216000,0.766704157157382)
--(axis cs:38592000,0.759413426089882)
--(axis cs:39984000,0.756567250794271)
--(axis cs:39984000,0.909119074445075)
--(axis cs:39984000,0.909119074445075)
--(axis cs:38592000,0.922568161865893)
--(axis cs:37216000,0.894949082662807)
--(axis cs:35840000,0.9183041737674)
--(axis cs:34464000,0.905820989529064)
--(axis cs:33088000,0.877292466628945)
--(axis cs:31712000,0.900834843887693)
--(axis cs:30336000,0.911420894145617)
--(axis cs:28960000,0.925784318951865)
--(axis cs:27584000,0.904363407804198)
--(axis cs:26192000,0.938982363315513)
--(axis cs:24816000,0.932446224290103)
--(axis cs:23440000,0.935065288170235)
--(axis cs:22064000,0.939391976424277)
--(axis cs:20688000,0.913265128085429)
--(axis cs:19312000,0.936485740225807)
--(axis cs:17936000,0.957325431393341)
--(axis cs:16560000,0.966895957988095)
--(axis cs:15184000,0.957051139519539)
--(axis cs:13808000,0.957470746534967)
--(axis cs:12416000,0.960157977622287)
--(axis cs:11040000,0.947543341444399)
--(axis cs:9664000,0.921692061383595)
--(axis cs:8288000,0.91769866745177)
--(axis cs:6912000,0.812575978577566)
--(axis cs:5536000,0.755541882141891)
--(axis cs:4160000,0.57493932283154)
--(axis cs:2784000,0.282731170366808)
--(axis cs:1408000,0.0869099962611728)
--(axis cs:32000,0)
--cycle;

\addplot [thick, C4, mark=*, mark size=0, mark options={solid}]
table {%
32000 0
1408000 0.0783685456490316
2784000 0.263974703383367
4160000 0.511290809614536
5536000 0.692769990380699
6912000 0.798321540781057
8288000 0.893603959410563
9664000 0.876369603802875
11040000 0.914433667541775
12416000 0.928664273818797
13808000 0.937465141311649
15184000 0.928676440185978
16560000 0.947057385693397
17936000 0.939860489355064
19312000 0.927334082202594
20688000 0.901735950302884
22064000 0.924347846468375
23440000 0.925355625531517
24816000 0.911657535128175
26192000 0.902601567821065
27584000 0.898840361075518
28960000 0.900323210245704
30336000 0.880041970996525
31712000 0.854107007358705
33088000 0.856026291879115
34464000 0.860270661406854
35840000 0.854880814373211
37216000 0.828323985500019
38592000 0.828786465467426
39984000 0.826819476697822
};

\path [draw=C1, fill=C1, opacity=0.2]
(axis cs:0,0)
--(axis cs:0,0)
--(axis cs:1600000,0)
--(axis cs:3200000,0.0444444444444444)
--(axis cs:4800000,0.377777777777778)
--(axis cs:6400000,0.466666666666667)
--(axis cs:8000000,0.544444444444444)
--(axis cs:9600000,0.744444444444444)
--(axis cs:11200000,0.755555555555556)
--(axis cs:12800000,0.5)
--(axis cs:14400000,0.577777777777778)
--(axis cs:16000000,0.7)
--(axis cs:17600000,0.733333333333333)
--(axis cs:19200000,0.711111111111111)
--(axis cs:20800000,0.8)
--(axis cs:22400000,0.822222222222222)
--(axis cs:24000000,0.755555555555556)
--(axis cs:25600000,0.766666666666667)
--(axis cs:27200000,0.844444444444445)
--(axis cs:28800000,0.844444444444445)
--(axis cs:30400000,0.755555555555556)
--(axis cs:32000000,0.9)
--(axis cs:33600000,0.833333333333333)
--(axis cs:35200000,0.822222222222222)
--(axis cs:36800000,0.9)
--(axis cs:38400000,0.877777777777778)
--(axis cs:40000000,0.922222222222222)
--(axis cs:40000000,1)
--(axis cs:38400000,0.955555555555555)
--(axis cs:36800000,1)
--(axis cs:35200000,0.966666666666667)
--(axis cs:33600000,0.988888888888889)
--(axis cs:32000000,0.988888888888889)
--(axis cs:30400000,0.977777777777778)
--(axis cs:28800000,0.988888888888889)
--(axis cs:27200000,0.933333333333333)
--(axis cs:25600000,0.966666666666667)
--(axis cs:24000000,0.933333333333333)
--(axis cs:22400000,0.96694444444444)
--(axis cs:20800000,0.966666666666667)
--(axis cs:19200000,0.922222222222222)
--(axis cs:17600000,0.966666666666667)
--(axis cs:16000000,0.888888888888889)
--(axis cs:14400000,0.877777777777778)
--(axis cs:12800000,0.811111111111111)
--(axis cs:11200000,0.966666666666667)
--(axis cs:9600000,0.933333333333333)
--(axis cs:8000000,0.811111111111111)
--(axis cs:6400000,0.833333333333333)
--(axis cs:4800000,0.6)
--(axis cs:3200000,0.255555555555556)
--(axis cs:1600000,0.0333333333333333)
--(axis cs:0,0)
--cycle;

\addplot [thick, C1, mark=*, mark size=0, mark options={solid}]
table {%
0 0
1600000 0
3200000 0.133333333333333
4800000 0.511111111111111
6400000 0.677777777777778
8000000 0.7
9600000 0.855555555555556
11200000 0.888888888888889
12800000 0.688888888888889
14400000 0.766666666666667
16000000 0.8
17600000 0.877777777777778
19200000 0.844444444444445
20800000 0.9
22400000 0.911111111111111
24000000 0.855555555555556
25600000 0.888888888888889
27200000 0.9
28800000 0.922222222222222
30400000 0.922222222222222
32000000 0.944444444444444
33600000 0.922222222222222
35200000 0.9
36800000 0.955555555555555
38400000 0.911111111111111
40000000 0.966666666666667
};

\addplot [thick, C2, mark=*, mark size=0, mark options={solid}]
table {%
0 0
1405200 0
2810400 0
4215600 0
5620800 0
7026000 0
8431200 0
9836400 0.02
11241600 0.02
12646800 0
14052000 0.07
15457200 0.07
16862400 0.09
18267600 0.08
19672800 0.08
21078000 0.09
22483200 0.17
23888400 0.14
25293600 0.17
26698800 0.2
28104000 0.26
29509200 0.18
30914400 0.23
32319600 0.29
33724800 0.27
35130000 0.25
36535200 0.27
37940400 0.33
39345600 0.31
};

\path [draw=C2, fill=C2, opacity=0.2]
% BBRL-PPO-dense
(axis cs:0,0)
--(axis cs:0,0)
--(axis cs:1405200,0)
--(axis cs:2810400,0)
--(axis cs:4215600,0)
--(axis cs:5620800,0)
--(axis cs:7026000,0)
--(axis cs:8431200,0)
--(axis cs:9836400,0)
--(axis cs:11241600,0)
--(axis cs:12646800,0)
--(axis cs:14052000,0.03)
--(axis cs:15457200,0.03)
--(axis cs:16862400,0.04)
--(axis cs:18267600,0.04)
--(axis cs:19672800,0.03)
--(axis cs:21078000,0.04)
--(axis cs:22483200,0.11)
--(axis cs:23888400,0.02)
--(axis cs:25293600,0.1)
--(axis cs:26698800,0.11)
--(axis cs:28104000,0.16)
--(axis cs:29509200,0.12)
--(axis cs:30914400,0.12)
--(axis cs:32319600,0.16)
--(axis cs:33724800,0.14)
--(axis cs:35130000,0.14)
--(axis cs:36535200,0.14)
--(axis cs:37940400,0.16)
--(axis cs:39345600,0.18)
--(axis cs:39345600,0.45)
--(axis cs:37940400,0.48)
--(axis cs:36535200,0.42)
--(axis cs:35130000,0.4)
--(axis cs:33724800,0.45)
--(axis cs:32319600,0.45)
--(axis cs:30914400,0.33)
--(axis cs:29509200,0.31)
--(axis cs:28104000,0.36)
--(axis cs:26698800,0.36)
--(axis cs:25293600,0.3)
--(axis cs:23888400,0.3)
--(axis cs:22483200,0.24)
--(axis cs:21078000,0.19)
--(axis cs:19672800,0.15)
--(axis cs:18267600,0.16)
--(axis cs:16862400,0.18)
--(axis cs:15457200,0.13)
--(axis cs:14052000,0.11)
--(axis cs:12646800,0.03)
--(axis cs:11241600,0.06)
--(axis cs:9836400,0.1)
--(axis cs:8431200,0.03)
--(axis cs:7026000,0.03)
--(axis cs:5620800,0.01)
--(axis cs:4215600,0)
--(axis cs:2810400,0)
--(axis cs:1405200,0)
--(axis cs:0,0)
--cycle;

\path [draw=C9, fill=C9, opacity=0.2]
(axis cs:32000,0)
--(axis cs:32000,0)
--(axis cs:1408000,0.00739769346474881)
--(axis cs:2784000,0.0106373002839909)
--(axis cs:4160000,0.0155479823583428)
--(axis cs:5536000,0.0229549040666607)
--(axis cs:6912000,0.0235877734838527)
--(axis cs:8288000,0.0311567873533867)
--(axis cs:9664000,0.0278272278537496)
--(axis cs:11040000,0.0369075214375936)
--(axis cs:12416000,0.0362778832552385)
--(axis cs:13808000,0.038350893509083)
--(axis cs:15184000,0.039076020909382)
--(axis cs:16560000,0.0452229900243051)
--(axis cs:17936000,0.0336153263527929)
--(axis cs:19312000,0.0403804865142632)
--(axis cs:20688000,0.033403670836089)
--(axis cs:22064000,0.0437874310489133)
--(axis cs:23440000,0.0461086089377064)
--(axis cs:24816000,0.0417985155926205)
--(axis cs:26192000,0.0449979652155313)
--(axis cs:27584000,0.0476323351942471)
--(axis cs:28960000,0.0509186615592193)
--(axis cs:30336000,0.057729944738985)
--(axis cs:31712000,0.0536379647612516)
--(axis cs:33088000,0.0577036078670841)
--(axis cs:34464000,0.0602775932807832)
--(axis cs:35840000,0.0581763436092858)
--(axis cs:37216000,0.0715009584815473)
--(axis cs:38592000,0.0797218162850232)
--(axis cs:38592000,0.144130354328308)
--(axis cs:38592000,0.144130354328308)
--(axis cs:37216000,0.138203133019802)
--(axis cs:35840000,0.120981957156012)
--(axis cs:34464000,0.11780364525105)
--(axis cs:33088000,0.100109662785585)
--(axis cs:31712000,0.105164062969023)
--(axis cs:30336000,0.0913552374947399)
--(axis cs:28960000,0.0866474891005595)
--(axis cs:27584000,0.0753134002848802)
--(axis cs:26192000,0.0661244807719177)
--(axis cs:24816000,0.0715251184080758)
--(axis cs:23440000,0.0664725882538657)
--(axis cs:22064000,0.0629505708527477)
--(axis cs:20688000,0.0562077410969471)
--(axis cs:19312000,0.0638896721751659)
--(axis cs:17936000,0.0494088115150066)
--(axis cs:16560000,0.0651396192973135)
--(axis cs:15184000,0.0534595821692715)
--(axis cs:13808000,0.05662890133403)
--(axis cs:12416000,0.0488032510425615)
--(axis cs:11040000,0.0557705704717284)
--(axis cs:9664000,0.0437661178400135)
--(axis cs:8288000,0.0492947462418009)
--(axis cs:6912000,0.0362603430635913)
--(axis cs:5536000,0.0323090330829951)
--(axis cs:4160000,0.0281906658220005)
--(axis cs:2784000,0.0185341883698532)
--(axis cs:1408000,0.0197281582278257)
--(axis cs:32000,0)
--cycle;

\addplot [thick, C9, mark=*, mark size=0, mark options={solid}]
table {%
32000 0
1408000 0.0127870448691096
2784000 0.0150055987911713
4160000 0.0213485783929655
5536000 0.0274885480507757
6912000 0.028576595965501
8288000 0.0406754187091681
9664000 0.0351401782683359
11040000 0.0468701744211427
12416000 0.0426974617533565
13808000 0.0457663066817507
15184000 0.044425927995989
16560000 0.0550487004668524
17936000 0.0439599343110962
19312000 0.0492501585469637
20688000 0.0440689287207225
22064000 0.0526768031985488
23440000 0.0555023654514538
24816000 0.0559427066978136
26192000 0.0584452392201698
27584000 0.062655008124692
28960000 0.0688920228500693
30336000 0.071108375383798
31712000 0.0772440238498992
33088000 0.0797819551666679
34464000 0.088982061125933
35840000 0.0917182040158085
37216000 0.105823580151453
38592000 0.114390953810025
};
\end{axis}

\end{tikzpicture}

%% file: tikz/legend_eval_FG_HP.tex
\begin{tikzpicture} 
    \begin{axis}[%
    hide axis,
    xmin=10,
    xmax=50,
    ymin=0,
    ymax=0.4,
    legend style={
        draw=white!15!black,
        legend cell align=left,
        legend columns=2, 
        legend style={
            draw=none,
            column sep=1ex,
            line width=1pt
        }
    },
    ]
    \addlegendimage{C1}
    \addlegendentry{Fancy Gym step-based};
    \addlegendimage{C4}
    \addlegendentry{Orbit step-based};
    \addlegendimage{C2}
    \addlegendentry{Fancy Gym BB};
    \addlegendimage{C9}
    \addlegendentry{Orbit BB};
    \end{axis}
\end{tikzpicture}

%% file: appendix.tex
\newpage
\onecolumn
\appendix

\subsection{Environment configuration}

\begin{figure}[!h]
    \centering
\begin{lstlisting}[label=lst:env-cfg,language=python]
@configclass
class ObjectTableSceneCfg(InteractiveSceneCfg):
    # robots: will be populated
    robot: ArticulationCfg = MISSING
    # end-effector sensor: will be populated
    ee_frame: FrameTransformerCfg = MISSING
    # target object: will be populated
    object: RigidObjectCfg = MISSING
    # Table
    table = AssetBaseCfg(...)
    # plane
    plane = AssetBaseCfg(...)
    # lights
    light = AssetBaseCfg(...)

@configclass
class CommandsCfg:
    ...

@configclass
class ActionsCfg:
    ...

@configclass
class ObservationsCfg:
    ...

@configclass
class RandomizationCfg:
    ...

@configclass
class RewardsCfg:
    ...

@configclass
class TerminationsCfg:
    ...

@configclass
class BoxPushingEnvCfg(RLTaskEnvCfg):

    # Scene settings
    scene: ObjectTableSceneCfg = ObjectTableSceneCfg(...)
    # Basic settings
    observations: ObservationsCfg = ObservationsCfg()
    actions: ActionsCfg = ActionsCfg()
    commands: CommandsCfg = CommandsCfg()
    # MDP settings
    rewards: RewardsCfg = RewardsCfg()
    terminations: TerminationsCfg = TerminationsCfg()
    randomization: RandomizationCfg = RandomizationCfg()

    def __post_init__(self):
    	# Sim parameters
        ...
        # PhysX parameters
        ...
\end{lstlisting}
\end{figure}

\newpage
\subsection{Hyper parameters}

\begin{table}[ht]
\centering
\caption{Hyperparameters for the box pushing experiments.}
\label{tab:boxpushing-HP}
\begin{adjustbox}{max width=\textwidth}
\begin{tabular}{lcccc}
                                 & step-based PPO Fancy Gym HP          & step-based PPO Orbit HP          & BBRL PPO Fancy Gym HP     & BBRL PPO Orbit HP     \\ 
\hline
\multicolumn{5}{l}{}                                                                                                  \\
number samples                   & 16000        & 98304        & 160          & 4096           \\
number parallel environments     & 160          & 4096          & n.a.         & 4096           \\
GAE $\lambda$                    & 0.95         & 0.95         & n.a.         & 0.95          \\
discount factor                  & 0.99         & 0.98         & n.a.         & 0.98          \\
\multicolumn{5}{l}{}                                                                                                  \\ 
\hline
\multicolumn{5}{l}{}                                                                                                  \\
epochs                           & 10           & 10           & 100          & 100           \\
learning rate                    & 1e-4         & 1e-4         & 1e-4         & 1e-4          \\
use critic                       & True         & True         & True         & True          \\
epochs critic                    & 10           & 5           & 100          & 100           \\
learning rate critic (and alpha) & 1e-4         & 1e-4         & 1e-4         & 1e-4          \\
number minibatches               & 40           & 4            & 1            & 1             \\
\multicolumn{5}{l}{}                                                                                                  \\ 
\hline
\multicolumn{5}{l}{}                                                                                                  \\
normalized observations          & True         & True         & False        & False         \\
normalized rewards               & True         & n.a.         & False        & n.a.         \\
observation clip                 & 10.0         & n.a.         & n.a.         & n.a.          \\
reward clip                      & 10.0         & n.a.         & n.a.         & n.a.          \\
critic clip                      & 0.2          & n.a.         & 0.2          & n.a.          \\
importance ratio clip            & 0.2          & n.a.         & 0.2          & n.a.          \\
\multicolumn{5}{l}{}                                                                                                  \\ 
\hline
\multicolumn{5}{l}{}                                                                                                  \\
hidden layers                    & {[}256, 256] & {[}256, 128, 64] & {[}128, 128] & {[}256, 128, 64]  \\
hidden layers critic             & {[}256, 256] & {[}256, 128, 64] & {[}32, 32]   & {[}256, 128, 64]    \\
hidden activation                & tanh         & elu         & tanh         & elu          \\
initial std                      & 1.0          & 1.0          & 1.0          & 1.0           \\
\multicolumn{5}{l}{}                                                                                                  \\ 
\hline
\multicolumn{5}{l}{}                                                                                                  \\
number basis functions           & n.a.         & n.a          & 5            & 5             \\
number zero basis                & n.a.         & n.a.         & 1            & n.a.          \\
weight scale                     & n.a.         & n.a.         & n.a.         & 0.3         
\end{tabular}
\end{adjustbox}
\end{table}

%% file: bare_conf.bbl
\begin{thebibliography}{10}

\bibitem{andrychowicz2017hindsight}
Marcin Andrychowicz, Filip Wolski, Alex Ray, Jonas Schneider, Rachel Fong, Peter Welinder, Bob McGrew, Josh Tobin, OpenAI Pieter~Abbeel, and Wojciech Zaremba.
\newblock Hindsight experience replay.
\newblock {\em Advances in neural information processing systems}, 30, 2017.

\bibitem{bartovs2021overview}
Michal Barto{\v{s}}, Vladim{\'\i}r Bulej, Martin Bohu{\v{s}}{\'\i}k, J{\'a}n Stan{\v{c}}ek, Vitalii Ivanov, and Peter Macek.
\newblock An overview of robot applications in automotive industry.
\newblock {\em Transportation Research Procedia}, 55:837--844, 2021.

\bibitem{benitti2012exploring}
Fabiane Barreto~Vavassori Benitti.
\newblock Exploring the educational potential of robotics in schools: A systematic review.
\newblock {\em Computers \& Education}, 58(3):978--988, 2012.

\bibitem{enayati2016haptics}
Nima Enayati, Elena De~Momi, and Giancarlo Ferrigno.
\newblock Haptics in robot-assisted surgery: Challenges and benefits.
\newblock {\em IEEE reviews in biomedical engineering}, 9:49--65, 2016.

\bibitem{haubold2020introducing}
Anne-Katrin Haubold, Lisa Obst, and Franziska Bielefeldt.
\newblock Introducing service robotics in inpatient geriatric care—a qualitative systematic review from a human resources perspective.
\newblock {\em Gruppe. Interaktion. Organisation. Zeitschrift F{\"u}r Angewandte Organisationspsychologie (GIO)}, 51(3):259--271, 2020.

\bibitem{korber2021comparing}
Marian K{\"o}rber, Johann Lange, Stephan Rediske, Simon Steinmann, and Roland Gl{\"u}ck.
\newblock Comparing popular simulation environments in the scope of robotics and reinforcement learning.
\newblock 2021.

\bibitem{lee2018development}
Carman~KM Lee.
\newblock Development of an industrial internet of things (iiot) based smart robotic warehouse management system.
\newblock In {\em CONF-IRM 2018 Proceedings. 43}, 2018.

\bibitem{mittal2024faq}
Mayank Mittal.
\newblock Orbit - frequently asked questions, 2024.
\newblock \url{https://isaac-orbit.github.io/orbit/source/refs/faq.html} [Accessed: (March 16, 2024)].

\bibitem{mittal2024known}
Mayank Mittal.
\newblock Orbit - known issues, 2024.
\newblock \url{https://isaac-orbit.github.io/orbit/source/refs/issues.html#non-determinism-in-physics-simulation} [Accessed: (March 07, 2024)].

\bibitem{mittal2023orbit}
Mayank Mittal, Calvin Yu, Qinxi Yu, Jingzhou Liu, Nikita Rudin, David Hoeller, Jia~Lin Yuan, Ritvik Singh, Yunrong Guo, Hammad Mazhar, Ajay Mandlekar, Buck Babich, Gavriel State, Marco Hutter, and Animesh Garg.
\newblock Orbit: A unified simulation framework for interactive robot learning environments.
\newblock {\em IEEE Robotics and Automation Letters}, 8(6):3740--3747, 2023.

\bibitem{morales2021survey}
Eduardo~F Morales, Rafael Murrieta-Cid, Israel Becerra, and Marco~A Esquivel-Basaldua.
\newblock A survey on deep learning and deep reinforcement learning in robotics with a tutorial on deep reinforcement learning.
\newblock {\em Intelligent Service Robotics}, 14(5):773--805, 2021.

\bibitem{otto2024fancy_gym}
Fabian Otto, Onur Celik, Dominik Roth, and Hongyi Zhou.
\newblock Fancy gym.

\bibitem{otto2023deep}
Fabian Otto, Onur Celik, Hongyi Zhou, Hanna Ziesche, Vien~Anh Ngo, and Gerhard Neumann.
\newblock Deep black-box reinforcement learning with movement primitives.
\newblock In {\em Conference on Robot Learning}, pages 1244--1265. PMLR, 2023.

\bibitem{otto2023mp3}
Fabian Otto, Hongyi Zhou, Onur Celik, Ge~Li, Rudolf Lioutikov, and Gerhard Neumann.
\newblock Mp3: Movement primitive-based (re-) planning policy.
\newblock {\em arXiv e-prints}, pages arXiv--2306, 2023.

\bibitem{paraschos2013probabilistic}
Alexandros Paraschos, Christian Daniel, Jan~R Peters, and Gerhard Neumann.
\newblock Probabilistic movement primitives.
\newblock {\em Advances in neural information processing systems}, 26, 2013.

\bibitem{plappert2018multi}
Matthias Plappert, Marcin Andrychowicz, Alex Ray, Bob McGrew, Bowen Baker, Glenn Powell, Jonas Schneider, Josh Tobin, Maciek Chociej, Peter Welinder, et~al.
\newblock Multi-goal reinforcement learning: Challenging robotics environments and request for research.
\newblock {\em arXiv e-prints}, pages arXiv--1802, 2018.

\bibitem{stulp2012reinforcement}
Freek Stulp, Evangelos~A Theodorou, and Stefan Schaal.
\newblock Reinforcement learning with sequences of motion primitives for robust manipulation.
\newblock {\em IEEE Transactions on robotics}, 28(6):1360--1370, 2012.

\bibitem{sutton2018reinforcement}
Richard~S Sutton and Andrew~G Barto.
\newblock {\em Reinforcement learning: An introduction}.
\newblock MIT press, 2018.

\bibitem{zhao2020sim}
Wenshuai Zhao, Jorge~Pe{\~n}a Queralta, and Tomi Westerlund.
\newblock Sim-to-real transfer in deep reinforcement learning for robotics: a survey.
\newblock In {\em 2020 IEEE symposium series on computational intelligence (SSCI)}, pages 737--744. IEEE, 2020.

\end{thebibliography}
